\newif\ifarxiv
\arxivtrue

\documentclass[letterpaper]{article} 
\usepackage{aaai2026}  
\usepackage{times}  
\usepackage{helvet}  
\usepackage{courier}  
\usepackage[hyphens]{url}  
\usepackage{graphicx} 
\usepackage{natbib}  
\usepackage{caption} 
\makeatletter
\newcommand{\leqnomode}{\tagsleft@true\let\veqno\@@leqno}
\makeatother

\usepackage{amssymb,amsthm,amsmath,bm,mathrsfs}
\usepackage{mathtools}
\usepackage{adjustbox}
\usepackage{enumitem}
\usepackage{etoc}
\usepackage{caption}
\usepackage{subcaption}
\usepackage{centernot}
\usepackage{complexity}
\usepackage{tabularx}   
\usepackage{multirow}   
\usepackage{array}      
\usepackage{booktabs}   
\usepackage{colortbl}   
\usepackage{xargs}      
\usepackage{tikz}
\usepackage{tikz-cd}
\usepackage{url}
\usepackage{makecell}   
\usepackage[ruled,linesnumbered,noend]{algorithm2e} 
\usepackage{lipsum}
\usepackage{listings}
\usepackage{soul}       
\usepackage{minitoc}    
\usepackage[textsize=scriptsize]{todonotes}
\usepackage{cleveref}

\usetikzlibrary{calc,positioning}
\setenumerate[1]{label={(\arabic*)}}

\newcommand{\tgi}{TGI}
\newcommand{\ptgi}{pTGI}
\newcommand{\btgi}{TGI$_C$}
\newcommand{\ogi}{OGI}
\newcommand{\sgi}{SGI}
\newcommand{\psgi}{pSGI}
\newcommand{\pogi}{pOGI}

\newclass{\PTIME}{PTIME}
\newclass{\EXPTIME}{EXPTIME}

\newcommand{\plansat}{\textsc{PlanSat}}
\newcommand{\planopt}{\textsc{PlanOpt}}

\newtheorem{example}{Example}
\newtheorem{theorem}{Theorem}

\newtheorem{corollary}[theorem]{Corollary}

\newtheorem{proposition}[theorem]{Proposition}
\theoremstyle{definition}
\theoremstyle{definition}
\newtheorem{definition}[theorem]{Definition}

\newcommand{\moose}{\textsc{Moose}}
\newcommand{\head}{\schemaF{actions}}
\newcommand{\stateBody}{\schemaF{stateCond}}
\newcommand{\goalBody}{\schemaF{goalCond}}
\newcommand{\rules}{\mathcal{R}}
\newcommand{\solution}{\pi}

\SetInd{0.5em}{0.5em}  

\newcommand{\stcp}[1]{{\algorithmFontSize{\tcp{#1}}}}
\newcommand{\fFont}[1]{\mathrm{#1}}
\newcommand{\cFont}[1]{\mathsf{#1}}

\SetKwInput{KwInput}{Input}
\SetKwInput{KwOutput}{Output}

\newcommand{\colorofcell}{blue}
\newcolumntype{Y}{>{\raggedleft\arraybackslash}X}
\newcommand{\header}[1]{\rotatebox[origin=l]{90}{\hspace*{-0.225cm} #1 }}
\newcommand{\first}[1]{\cellcolor{\colorofcell!40}{\textbf{#1}}}

\definecolor{caribbeangreen}{rgb}{0.0, 0.8, 0.6}
\definecolor{brilliantlavender}{rgb}{0.96, 0.73, 1.0}
\definecolor{amethyst}{rgb}{0.6, 0.4, 0.8}
\definecolor{ao(english)}{rgb}{0.0, 0.5, 0.0}
\definecolor{arylideyellow}{rgb}{0.91, 0.84, 0.42}
\definecolor{asparagus}{rgb}{0.53, 0.66, 0.42}
\definecolor{aquamarine}{rgb}{0.5, 1.0, 0.83}
\definecolor{babyblue}{rgb}{0.54, 0.81, 0.94}
\definecolor{rosewood}{rgb}{0.4, 0.0, 0.04}
\definecolor{oldmauve}{rgb}{0.4, 0.19, 0.28}
\definecolor{myrtle}{rgb}{0.13, 0.26, 0.12}
\definecolor{magenta(dye)}{rgb}{0.79, 0.08, 0.48}

\definecolor{plta}{rgb}{0.12, 0.47, 0.71}
\definecolor{pltb}{rgb}{   1, 0.5, 0.05}
\definecolor{pltc}{rgb}{0.17, 0.63, 0.17}
\definecolor{pltd}{rgb}{0.84, 0.15, 0.16}

\def\N{\mathbb{N}}
\def\R{\mathbb{R}}

\def\a{\alpha}
\def\b{\beta}

\renewcommand{\phi}{\varphi}

\def\la{\leftarrow}

\newcommand{\abs}[1]{\left| #1 \right|}

\newcommand{\gen}[1]{\left< #1 \right>}

\newcommand{\set}[1]{\left\{ #1 \right\}}
\newcommand{\seta}[1]{\{ #1 \}}

\DeclareMathOperator*{\concatsmall}{%
  {\Vert}%
}

\newcommand{\domain}{\mathcal{D}}
\newcommand{\problem}{\mathbf{P}}
\newcommand{\problems}{\mathscr{P}}
\newcommand{\trainProblems}{\problems_{\text{train}}}
\newcommand{\testProblems}{\problems_{\text{test}}}

\newcommand{\gproblem}{\mathbf{GP}}
\newcommand{\GP}{\gproblem}

\newcommand{\gptuple}{\gen{\trainProblems, \testProblems}}
\newcommand{\objects}{{O}}
\newcommand{\constants}{\mathcal{C}}
\newcommand{\predicates}{\mathcal{P}}
\newcommand{\functions}{\mathcal{F}}
\newcommand{\schemata}{\mathcal{A}}

\newcommand{\mi}[1]{\ensuremath{\mathit{#1}}}

\newcommand{\schemaF}[1]{\mathit{#1}}
\newcommand{\var}{\schemaF{var}}
\newcommand{\pre}{\schemaF{pre}}
\newcommand{\add}{\schemaF{add}}
\newcommand{\del}{\schemaF{del}}
\newcommand{\numEff}{\schemaF{num\_eff}}

\renewcommand{\succ}{\fFont{succ}}
\newcommand{\regr}{\fFont{regr}}
\newcommand{\grounding}{\fFont{grounding}}
\newcommand{\gplan}{\pi}
\newcommand{\plan}{\vec{\alpha}}
\newcommand{\arity}{\mathrm{ar}}

\newcommand{\loom}{\textsc{Loom}}
\newcommand{\prp}{\textsc{PRP}}
\newcommand{\slearner}{\textsc{SLearn}}
\newcommand{\slearnerA}{\textsc{SLearn-0}}
\newcommand{\slearnerB}{\textsc{SLearn-1}}
\newcommand{\slearnerC}{\textsc{SLearn-2}}
\newcommand{\blind}{Blind}
\newcommand{\lmcut}{\textsc{LMcut}}
\newcommand{\enhsp}{\textsc{Enhsp}}
\newcommand{\enhspMQ}{{$\textsc{M}(3h\!\concatsmall\!3n)$}}
\newcommand{\enhspHMRPHJ}{\textsc{mrp+hj}}
\newcommand{\lama}{\textsc{Lama}}
\newcommand{\symk}{\textsc{SymK}}
\newcommand{\scorpion}{\textsc{Scorpion}}

\newcommand{\algorithmFontSize}{
}

\title{Satisficing and Optimal Generalised Planning via Goal Regression\\(Extended Version)}
\author{
  Dillon Z. Chen$^{1,2,3}$,
  Till Hofmann$^4$,
  Toryn Q. Klassen$^{1,3}$,
  Sheila A. McIlraith$^{1,3}$
}
\affiliations{
  \begin{tabular}{c c}
    $^1$Vector Institute, Toronto, Canada &
    $^2$LAAS-CNRS, University of Toulouse, France \\
    $^3$University of Toronto, Canada &
    $^4$RWTH Aachen University, Germany
  \end{tabular}
}

\begin{document}


\ifarxiv
\nocopyright
\fi


\maketitle
\begin{abstract}
  Generalised planning (GP) refers to the task of synthesising programs that solve families of related planning problems.
  We introduce a novel, yet simple method for GP: given a set of training problems, for each problem, compute an optimal plan for each goal atom in some order, perform goal regression on the resulting plans, and lift the corresponding outputs to obtain a set of first-order \emph{Condition} $\rightarrow$ \emph{Actions} rules.
  The rules collectively constitute a generalised plan that can be executed as is or alternatively be used to prune the planning search space.
  We formalise and prove the conditions under which our method is guaranteed to learn valid generalised plans and state space pruning axioms for search.
  Experiments demonstrate significant improvements over state-of-the-art (generalised) planners with respect to the 3 metrics of synthesis cost, planning coverage, and solution quality on various classical and numeric planning domains.
\end{abstract}

\ifarxiv\else
\begin{links}
  \link{Code}{https://github.com/dillonzchen/moose}
  \link{Dataset}{https://github.com/dillonzchen/moose-dataset}
  \link{Extended version}{TODO}
\end{links}
\fi

\section{Introduction}
Generalised planning (GP) aims to compute generalised plans: programs that solve families of related planning problems.
A grand goal of GP is to amortise synthesis costs by solving families of planning problems faster than general-purpose planners that solve each problem individually.
Indeed, there exists several planning domains that are computationally easy to solve and exhibit satisficing policies, such as variants of the package delivery domain~\cite{helmert.2003}.
In the real world, UPS$^{\text{TM}}$ delivered over 20 million packages daily across over 200 countries and territories in 2024~\cite{ups.2025}.
However, state-of-the-art, general-purpose planners struggle to scale up to a simplified version of the delivery problem with 100 packages~\cite{taitler.etal.2024}.

\begin{figure}
  \includegraphics[width=\columnwidth]{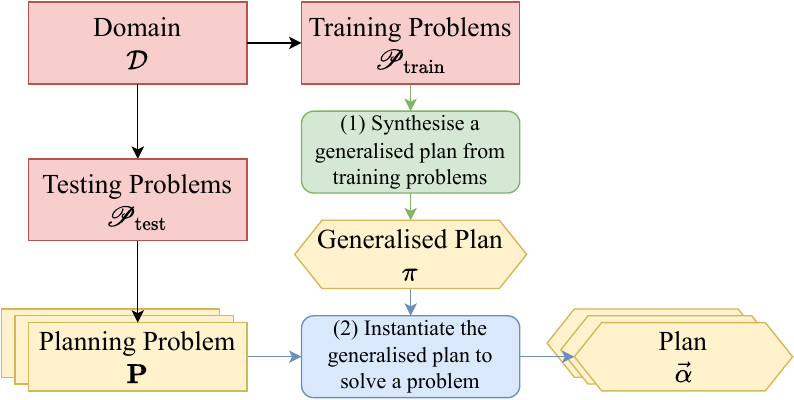}
  \caption{
    A common GP setup consisting of a planning domain $\domain$, and a set of training problems $\trainProblems$ and testing problems $\testProblems$.
    A generalised planner consists of two modules:
    (1) synthesis and
    (2) instantiation.
    See text for details.
  }
  \label{fig:gp}
\end{figure}

We consider the GP problem that consists of a planning domain $\domain$, and a set of training problems $\trainProblems$ and testing problems $\testProblems$ drawn from $\domain$, as depicted in \Cref{fig:gp}.
A generalised planner synthesises a generalised plan from $\domain$ and $\trainProblems$ that solves problems in $\testProblems$.
Metrics for evaluating the effectiveness of a generalised planner~\cite{srivastava.etal.2011a} include the resources it takes to synthesise a generalised plan (\emph{synthesis cost}), the time it takes to instantiate generalised plans on unseen problems (\emph{instantiation cost}), and the quality of instantiated plans (\emph{solution quality}).

In this paper, we introduce a new generalised planner that draws upon insights and long-standing ideas of \emph{goal regression} from the knowledge representation community and \emph{problem relaxation} from the planning community to advance the state of the art in GP over the three aforementioned metrics.
Our approach consists of an efficient three step process of (1) solving for each goal of each problem in $\trainProblems$ optimally in an arbitrary order, (2) performing goal regression~\cite{fikes.etal.1972,waldinger.1977,lozano.etal.1984,reiter.1991,reiter.2001} over the resulting plans, and (3) lifting the corresponding partial-state, macro-action pairs into sets of first-order rules constituting a generalised plan.
We treat planning states as databases~\cite{correa.etal.2020} and lifted rules as queries, and correspondingly employ database algorithms for instantiating generalised plans quickly on problems in $\testProblems$.
We also leverage derived predicates and axioms to encode learned rules for state space pruning in search.

\begin{figure*}[t]
  \centering
  \includegraphics[width=\textwidth]{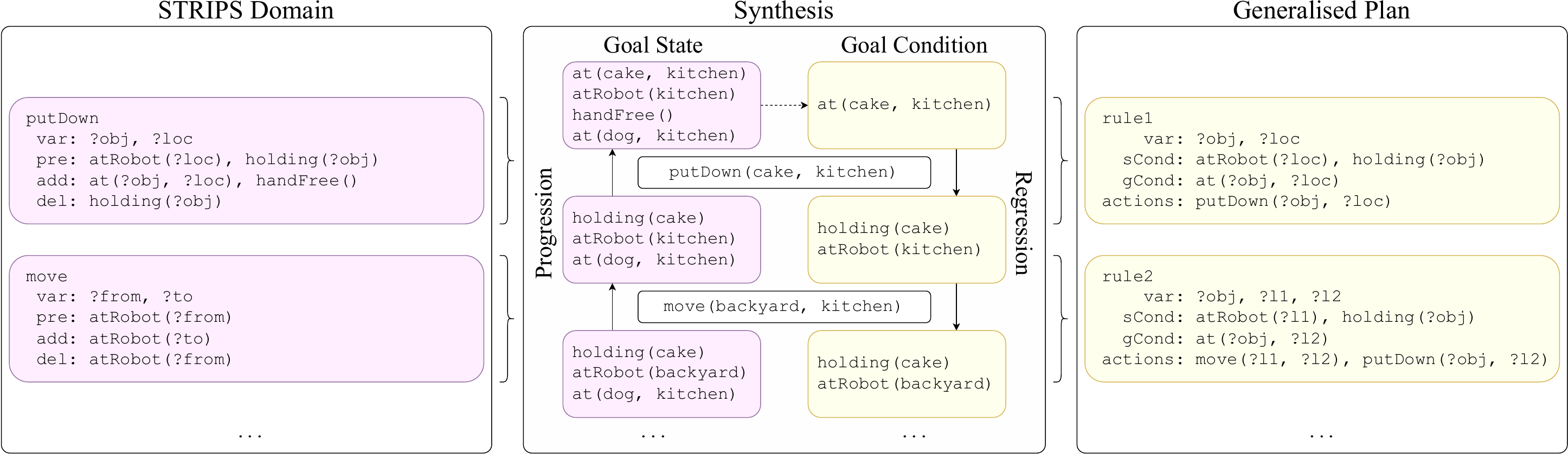}
  \caption{
    \textbf{Left:} a simplified STRIPS transportation domain.
    \textbf{Middle:} state progression (purple) and goal regression (yellow) via a \texttt{putDown} action.
    \textbf{Right:} the generalised plan created by lifting the regressed states, goal condition, and plan actions.
  }
  \label{fig:moose}
\end{figure*}

We formalise the conditions under which our approach learns sound and complete generalised plans, and under which the encoding of learned rules as axioms gives rise to provably optimal plans when combined with optimal search planners.
We manifest our contributions in the \moose{} planner\ifarxiv\footnote{Available at \url{https://github.com/dillonzchen/moose}}\fi and conduct experiments with \moose{} on classical and numeric planning domains and satisficing and optimal planning settings.
We observe that \moose{} outperforms state-of-the-art baseline planners by large margins on Easy-to-Solve, Hard-to-Optimise (ESHO) planning domains --- \P-time solvable and \NP-hard to solve optimally domains.

\noindent We summarily list our contributions as follows.
\begin{itemize}[leftmargin=*]
  \item We introduce algorithms for synthesising and instantiating generalised plans that employ goal regression, problem relaxation, and first-order query techniques.
  \item We formalise the conditions under which our approach is sound and complete for satisficing and optimal planning.
  \item We conduct experiments and demonstrate the effectiveness of our algorithms in satisficing and optimal planning over various classical and numeric planning domains.
\end{itemize}

\section{Planning Background and Notation}
We adopt standard notation for representing planning problems via the STRIPS fragment of the Planning Domain Definition Language (PDDL)~\cite{mcdermott.etal.1998,haslum.etal.2019}.
Our approach also handles a fragment of numeric planning but we defer such definitions to \Cref{app:sec:numeric-background}.

\newcommand{\defn}[1]{\emph{#1}}

\paragraph{Mathematical Notation}
Let $\N$/$\N_0$ denote the natural numbers excluding/including 0,
$\vec{\a}$ denote an ordered sequence of items,
$\vec{\a}_i$ denote the $i$th element of $\vec{\a}$,
$\vec{\a}_{[i:]}$ all elements from the $i$th element onwards in $\vec{\a}$ inclusive,
and $\abs{s}$ the size of a set or length of a sequence $s$.

\paragraph{STRIPS Planning}
A planning problem is represented by two components: a domain, consisting of lifted predicates and action schemata describing the action theory, and a problem specification, consisting of a finite set of objects, an initial state, and a goal condition.

A \defn{planning domain} is a tuple $\domain = \gen{\predicates, \constants, \schemata}$, where $\predicates$ is a set of predicates, $\constants$ a set of constant objects, and $\schemata$ a set of action schemata.
A predicate $p \in \predicates$ has a set of argument terms $x_1, \ldots, x_{n_p}$ where $n_p \in \N_0$ depends on $p$.
An action schema $a \in \schemata$ is a tuple
$
a = \gen{\var(a), \pre(a), \add(a), \del(a)}
$
where $\var(a)$ is a set of parameter variables, and preconditions $\pre(a)$, add $\add(a)$ and delete $\del(a)$ effects are finite sets of predicates from $\predicates$ with arguments instantiated with variables or objects from $\var(a) \cup \constants$.

\begin{example}[Transportation Domain]\label{ex:transportdomain}
  We can model a simplified transportation domain where an agent can transport items between locations in STRIPS, partially depicted in \Cref{fig:moose}, as follows.
  We define a domain $\domain = \gen{\predicates, \constants, \schemata}$ with
  $\predicates = \{ \mi{at}(\mi{?x}, \mi{?y}), \allowbreak \mi{atRobot}(\mi{?x}), \allowbreak \mi{handFree}(), \allowbreak \mi{holding}(\mi{?x}) \}$,
  $\constants = \emptyset$, and $\schemata$ containing the three action schemata:
  \begin{align*}
    \shortintertext{$\mi{putDown}$}
    \var &= \{ \mi{?obj}, \mi{?loc} \}
    \\
    \pre &= \{ \mi{atRobot}(\mi{?loc}), \mi{holding}(\mi{?obj})\}
    \\
    \add &= \{ \mi{at}(\mi{?obj}, \mi{?loc}), \mi{handFree}()\}
    \\
    \del &= \{ \mi{holding}(\mi{?obj})\}
    \shortintertext{$\mi{move}$}
    \var &= \{\mi{?from}, \mi{?to}\}
    \\
    \pre &= \{\mi{atRobot}(\mi{?from})\}
    \\
    \add &= \{\mi{atRobot}(\mi{?to})\}
    \\
    \del &= \{\mi{atRobot}(\mi{?from})\}
    \shortintertext{$\mi{pickUp}$}
    \var &= \{ \mi{?obj}, \mi{?loc} \}
    \\
    \pre &= \{ \mi{atRobot}(\mi{?loc}), \mi{at}(\mi{?obj}, \mi{?loc}), \mi{handFree}()\}
    \\
    \add &= \{ \mi{holding}(\mi{?obj})\}
    \\
    \del &= \{ \mi{at}(\mi{?obj}, \mi{?loc}), \mi{handFree}()\}
  \end{align*}
\end{example}

A \defn{planning problem} is a tuple $\problem = \gen{\domain, s_0, g, \objects}$ with $\domain$ a planning domain, $s_0$ the initial state, $g$ the goal condition, and $\objects \supseteq \constants$ a finite set of objects.
A (ground) atom is a predicate whose argument terms are all instantiated with objects.
A state in a planning problem is a set of atoms and operate under the closed world assumption: any atom not in a state is presumed false.
The initial state $s_0$ and goal condition $g$ are both sets of atoms.
A state $s$ is a goal state if $g \subseteq s$.
We say that the problem $\problem$ belongs to the domain $\domain$.
A (ground) action is an action schema $a$ where each parameter term is instantiated with an object, denoted $a(o_1, \ldots, o_n)$ with $o_1,\ldots,o_n \in \objects$.
An action $a$ is applicable in a state $s$ if $\pre(a) \subseteq s$, in which case we define the \defn{successor}
\begin{align*}
  \succ(s, a) = (s \setminus \del(a)) \cup \add(a).
\end{align*}
Otherwise, $a$ is not applicable in $s$ and $\succ(s, a) = \bot$.

A \defn{plan} for a planning problem is a finite sequence of actions $\plan = a_1, \ldots, a_n$ where $s_{i}=\succ(s_{i-1}, a_{i})\not=\bot$ for $i = 1, \ldots, n$ and $s_n$ is a goal state.
We overload the notation of successor for sequences of actions with $\succ(s, \plan) = s_n$ as if $s = s_0$.
The length of a plan $\plan$ is its number of actions.
A problem $\problem$ is solvable if a plan exists for $\problem$.
Satisficing (resp.\ optimal) planning refers to the task of finding any plan (resp.\ any plan with the lowest length) for $\problem$.

\begin{example}[Transportation Problem]\label{eg:prob}
  We model a transportation problem where a robot and a dog are in the kitchen, and a cake that is in the backyard has to be brought into the kitchen as a STRIPS problem $\problem = \gen{\domain, s_0, g, \objects}$
  where $\domain$ is the transportation domain in \Cref{ex:transportdomain},
  $s_0 = \{\mi{at}(\mi{cake}, \mi{backyard}), \allowbreak \mi{at}(\mi{dog}, \mi{kitchen}), \allowbreak \mi{atRobot}(\mi{kitchen}), \allowbreak \mi{handFree}()\}$,
  $g=\{\mi{at}(\mi{cake}, \mi{kitchen})\}$, and $\objects = \{\mi{backyard}, \mi{cake}, \mi{dog}, \mi{kitchen}\}$.
  A plan is given by $\plan =
  \mi{move(kitchen, backyard)}, \allowbreak
  \mi{pickUp(cake, backyard)}, \allowbreak
  \mi{move(backyard, kitchen)}, \allowbreak
  \mi{putDown(cake, kitchen)}$.
  The problem and plan are partially depicted in \Cref{fig:moose}.
\end{example}

We introduce some additional notational shorthands to be used later.
Let $\problem[\omega]$ denote the $\omega$ component of a problem $\problem$; e.g., $\problem[s_0]$ is the initial state of $\problem$.
Next, given a state $s$ and set of atoms $g'$ for a problem $\problem$, let
$\problem_{s} = \gen{\domain, s, \problem[g], \problem[\objects]}$
denote the same problem with the initial state replaced with $s$, and
$\problem_{s, g'} = \gen{\domain, s, g', \problem[\objects]}$
denote the same problem with the initial state and goal replaced with $s$ and $g'$.

\paragraph{Goal Regression}
Goal regression refers to the computation of the preimage of a goal under a sequence of actions via regression rewriting.
Goal regression computes the minimal set of goal relevant atoms, and has been used for heuristic synthesis~\cite{bonet.geffner.2001,scala.etal.2016}, plan monitoring~\cite{fritz.mcilraith.2007}, policy synthesis for lifted Markov decision processes~\cite{gretton.thiebaux.2004,sanner.boutilier.2009}, nondeterministic planning~\cite{muise.etal.2012,muise.etal.2024}, symbolic search~\cite{pang.holte.2011,alcazar.etal.2013,speck.etal.2025}, numeric planning~\cite{illanes.mcilraith.2017}, generating macro-actions~\cite{hofmann.etal.2020}, GP~\cite{illanes.mcilraith.2019,yang.etal.2022}, and embodied AI~\cite{kaelbling.lozano.2011,xu.etal.2019,liu.etal.2025}.

A set of atoms $g$ is regressable over an action $a$ if $\add(a) \cap g \not= \emptyset$ and $\del(a) \cap g = \emptyset$,
in which case we define the \defn{regression}
\begin{align*}
  \regr(g, a) = (g \setminus \add(a)) \cup \pre(a).
\end{align*}
Otherwise, $g$ is not regressable over $a$ and $\regr(g, a) = \bot$.

\subsection*{Problem Statement: Generalised Planning}
We introduce the generalised planning (GP) problem as a set of planning problems sharing the same domain.
We describe the variant involving a set of training problems as seen commonly in recent GP works (e.g. \cite{frances.etal.2021,drexler.etal.2022,yang.etal.2022}).

A \defn{generalised planning problem} is a tuple $\gproblem = \gptuple$ where $\trainProblems$ (resp. $\testProblems$) is a finite (resp.\ possibly infinite) set of problems belonging to the same domain $\domain$.
A \defn{generalised plan} $\pi$ for a GP problem is a programmatic plan that
is synthesised from the domain $\domain$ and $\trainProblems$, and
can be instantiated on any planning problem $\problem \in \testProblems$ to return a valid plan $\pi(\problem) = \plan$ if a plan exists for $\problem$, or otherwise determine that no plan exists for $\problem$.
Examples of programmatic plans include finite state controllers~\cite{bonet.etal.2009,bonet.etal.2010,hu.giacomo.2011,aguas.etal.2018}, policies derived from lifted rules~\cite{srivastava.etal.2011,illanes.mcilraith.2019,frances.etal.2021} and general-purpose programs~\cite{levesque.2005,srivastava.etal.2008,segoviaaguas.etal.2024,silver.etal.2024}.

\section{Generalised Planning via Goal Regression}
We name our generalised plans as \moose{} programs.
\moose{} programs are found in a two-step process as described in \Cref{ssec:learning}:
(a) decompose the set of training problems $\trainProblems$ into smaller problems constituting singleton goal conditions and generate optimal plans for each in order, and
(b) apply goal regression from the singleton goals using the order of the optimal plans found in (a) to generate a set of lifted rules.
\moose{} programs can be instantiated into a plan for a problem by deriving an action from the rule set at every state until the goal is reached (\Cref{ssec:satisficing}), or used to guide search for optimal planning (\Cref{ssec:optimal}).

\moose{} programs are sets of lifted rules which indicate an action or macro action to execute, conditioned on a partial state and a goal that has not yet been achieved.
A distinct feature of such rules is that the antecedent of a single rule compactly captures a set of states.
Lifted rules can then be grounded on states if their antecedent condition is satisfied.
Our rules are similar to existing lifted rules~\cite{khardon.1999,illanes.mcilraith.2019,yang.etal.2022} with the extension that we may now have macro actions in rule heads.
Furthermore, each rule has an associated precedence value that determines its execution priority as is common in logic programming.
\Cref{fig:moose} illustrates the synthesis procedure and structure of \moose{} programs.

\begin{definition}[\moose{} Rule]\label{defn:rule}
  Let $\gproblem = \gptuple$ be a GP problem.
  A \moose{} rule $r$ is a tuple
  \[ r = \langle \var(r), \stateBody(r), \goalBody(r), \head(r) \rangle \]
  where
  $\var(r)$ is a finite set of free variables,
  $\stateBody(r)$ and $\goalBody(r)$ are finite sets of predicates instantiated with terms in $\var(r)$, and
  $\head(r)$ is a finite sequence of action schemata instantiated with terms in $\var(r)$.
\end{definition}

\begin{definition}[Grounding]
  Let $r$ be a \moose{} rule, $\problem$ be a problem and $s$ a state in the state space of $\problem$.
  A grounding of $r$ in $s$ is an assignment of objects to variables $f: \var(r) \to \problem[\objects]$ such that $\stateBody(r)|_{f} \subseteq s$ and $\goalBody(r)|_{f} \subseteq (\problem[g] \setminus s)$, the set of goal atoms not yet achieved.
  The $|_{f}$ notation denotes replacing every occurrence of a free variable term with the corresponding object in $f$.
  In the case that a grounding $f$ exists, we define the nondeterministic function
  $\grounding(r, s, \problem[g]) = \head(r)|_{f},$
  where $f$ is some grounding.
  Otherwise, $\grounding(r, s, \problem[g])$ $=$ $\bot$.
\end{definition}

\begin{definition}[\moose{} Program]
  A \moose{} program $\solution$ is a set of \moose{} rules $\rules$ and a function $\rules \to \N$ representing a precedence ranking on the rules for execution.
\end{definition}

As to be described later, if several rules are applicable in a given state, the rule with the lowest precedence ranking with ties broken arbitrarily is chosen for execution.
Relatedly, \citet{yang.etal.2022} specify a total order on policy rules, whereas \moose{} specifies more relaxed partial order.
Next, we define lifting of a ground plan and set of atoms to a set of quantified actions and predicates.
Lifting will be used in the synthesis module to generate reusable rules.
\begin{definition}[Lifting]\label{def:lifting}
  Let $s$ and $g$ be finite sets of ground atoms and $\plan = a_1, \ldots, a_m$ a sequence of ground actions.
  Let $o_1, \ldots, o_q$ be the union of all objects from the atoms and actions that are not in $\constants$.
  Next we define the set of free variable terms $\var = \set{v_1, \ldots, v_q}$
  and lift each action and atom by replacing each constant $o_i$
  with its corresponding free variable $v_i$ in $\var$.
  We denote
  \begin{align*}
    \fFont{lift}(s, g, \plan) = \gen{\var, s', g', \plan'} \label{eqn:lift}
  \end{align*}
  with $\plan'$ the sequence of ground actions lifted by variables in $\var$, and similarly for $s'$ and $g'$ the sets of lifted atoms.
\end{definition}

\subsection{Synthesising \moose{} Programs}\label{ssec:learning}
\Cref{alg:learn} summaries the main \moose{} program synthesis procedure.
The input is a set of unlabelled training problems and a number $n_p$ representing the effort spent on extracting information from a single problem.
The main idea is that the problem is relaxed by decoupling the goals and greedily solving them optimally and in order.
Each resulting plan regresses the corresponding singleton goal, and the regressed goal and plan is then lifted into a lifted macro action rule.

\begin{algorithm}[t]
  \algorithmFontSize
  \DontPrintSemicolon

  \caption{\moose{} Program Synthesis}
  \label{alg:learn}
  \KwInput{
    Training problems $\trainProblems = \seta{\problem^{(1)}, \ldots, \problem^{(n_t)}}$, and number of goal permutations $n_p \in \N$ (default: 3).
  }
  \KwOutput{
    \moose{} program $\solution$.
  }
  $\solution \la \emptyset$ \label{line:learn:empty} \\
  \For{$i=1, \ldots, n_t$}{ \label{line:learn:instances}
    $n_g \la |\problem^{(i)}[g]|$ \\
    \For{$j=1, \ldots, \min(n_p, n_g!)$}{ \label{line:learn:permutations}
      $s \la \problem^{(i)}[s_0]$ ; 
      $\vec{g} \la \fFont{newPermutation}(\problem^{(i)}[g])$ \\
      \For{$k=1, \ldots, n_g$}{ \label{line:learn:goal}
        $g' \la \set{\vec{g}_k}$ \label{line:learn:singleton} \\
        $\plan \la \fFont{optimalPlan}(\problem^{(i)}_{s, g'})$ \label{line:learn:opt} \\
        \lIf{$\plan = \bot$}{
          \textbf{continue} \label{line:learn:deadend}
        }
        $\solution \la \solution \cup \fFont{extractRules}(\plan, g')$  \stcp{Alg.~\ref{alg:extract-rules}} \label{line:learn:extract}
        $s \la \succ(s, \plan)$ \label{line:learn:progress} \\
      }
    }
  }
  \Return{$\solution$}
\end{algorithm}

\begin{algorithm}[t]
  \algorithmFontSize
  \caption{Rule Extraction Routine}
  \label{alg:extract-rules}
  \KwInput{
    Sequence of actions $\plan$ and set of atoms $g$.
  }
  \KwOutput{
    \moose{} rules with precedence values $\solution$.
  }
  $\solution \la \emptyset$; $s \la g$ \label{line:extract:ug} \\
  \For{$i=\abs{\plan}, \ldots, 1$}{ \label{line:extract:regress1}
    $s \la \regr(s, \plan_i)$ ;
    $r \la \fFont{lift}(s, g, \plan_{[i:]})$ \label{line:extract:lift} \\
    $\solution \la \solution \cup \set{(r, \abs{\plan} - i + 1)}$ \label{line:extract:precedence} \label{line:extract:add} \\
  }
  \Return{$\solution$}
\end{algorithm}

\begin{algorithm}[t]
  \algorithmFontSize
  \DontPrintSemicolon

  \newcommand{\tmpplan}{\vec{\b}}

  \caption{\moose{} Program Instantiation}
  \label{alg:plan}
  \KwInput{
    A planning problem $\problem$ and \moose{} program $\solution$.
  }
  \KwOutput{
    A plan $\plan$ and $\cFont{success}$ or $\cFont{failure}$ status.
  }
  $s \la \problem[s_0]$; 
  $\plan \la []$  \stcp{empty sequence}
  \While{$\problem[g] \not\subseteq s$}{ \label{line:plan:main}
    $\tmpplan \la \bot$ \\
    \For{$r \in \solution$ in ascending precedence values}{ \label{line:plan:query1}
      $\tmpplan \la \grounding(r, s, \problem[g])$ \\
      \lIf{$\tmpplan \not= \bot$}{\textbf{break}} \label{line:plan:query2}
    }
    \lIf{$\tmpplan = \bot$ or detected cycle}{
      \Return{$\plan, \cFont{failure}$} \label{line:plan:failure}
    }
    $\plan \la \plan ; \tmpplan$ \label{line:plan:exec1}  \stcp{sequence concatenation}
    $s \la \succ(s, \tmpplan)$ \label{line:plan:exec2} \\
  }
  \Return{$\plan, \cFont{success}$}
\end{algorithm}

The main algorithm gradually builds from an empty rule set (\Cref{line:learn:empty}) by iterating over all training problems (\Cref{line:learn:instances}) and the specified number $n_p$ of goal orderings (\Cref{line:learn:permutations}).
For each goal ordering and training problem, \moose{} finds plans via an optimal planner (\Cref{line:learn:opt}) with the singleton goals (\Cref{line:learn:singleton}) in order (\Cref{line:learn:goal}), while progressing the current state along the way (\Cref{line:learn:progress}).
If no plan exists, i.e.\ if the problem is unsolvable with the current state and singleton goal pair, no rules are extracted and the state is not progressed (\Cref{line:learn:deadend}).
Otherwise, if a plan exists, then we extract rules from the plan and add them to the incumbent plan (\Cref{line:learn:extract}) as described in \Cref{alg:extract-rules}.
It begins by initialising the to-be-regressed state $s$ by the goal (\Cref{line:extract:ug}).
Next, it regresses $s$ in reverse order of the plan $\plan$ and then lifts the corresponding regressed state $s$, goal $g$, and suffix of the plan into a rule $r$ (\Crefrange{line:extract:regress1}{line:extract:lift}).
Then we append the rule alongside its cost-to-go from the partial state $s$ to goal $g$ under the plan suffix as its precedence value (\Cref{line:extract:add}).

\begin{example}[Transportation Program Synthesis]
  We illustrate Lines 8 to 10 of \Cref{alg:learn} with our running transportation example from \Cref{eg:prob} as a training problem.
  Note that the problem already has a singleton goal $g = \{\textit{at(cake, kitchen)}\}$.
  The plan $\plan$ from the example is the only optimal plan and thus is the output of Line 8.
  Since a plan exists for the problem, Line 9 does nothing.
  Line 10 triggers \Cref{alg:extract-rules} which begins by setting $s$ to $g$.
  The first regressed state under the final action in the plan
  $a=\mi{putDown}(\mi{cake}, \mi{kitchen})$ is $\regr(s, a)=\{\mi{atRobot}(\mi{kitchen}), \mi{holding}(\mi{cake}) \}$.
  Note that the fact $\mi{at}(\mi{dog}, \mi{kitchen})$ is ignored during regression, indicating that it is irrelevant towards the goal.

  The regressed state and the singleton goal is lifted to construct the following rule
  \begin{align*}
    \var &= \{ \mi{?obj}, \mi{?loc} \}
    \\
    \mi{stateCond} &= \{ \mi{atRobot}(\mi{?loc}), \mi{holding}(\mi{?obj})\}
    \\
    \mi{goalCond} &= \{ \mi{at}(\mi{?obj}, \mi{?loc})\}
    \\
    \mi{actions} &= \mi{putDown}(\mi{?obj}, \mi{?loc})
  \end{align*}

  Repeating the procedure again under the penultimate action in the plan gives us the next regressed state
  $\{ \mi{atRobot}(\mi{backyard}), \mi{holding}(\mi{cake}) \}$
  which is lifted to construct the rule
  \begin{align*}
    \var &= \{ \mi{?obj}, \mi{?l1}, \mi{?l2} \}
    \\
    \mi{stateCond} &= \{ \mi{atRobot}(\mi{?l1}), \mi{holding}(\mi{?obj})\}
    \\
    \mi{goalCond} &= \{ \mi{at}(\mi{?obj}, \mi{?l2})\}
    \\
    \mi{actions} &= \mi{move}(\mi{?l1}, \mi{?l2}), \mi{putDown}(\mi{?obj}, \mi{?l2})
  \end{align*}
  The procedure is repeated two more times as $\plan$ has four actions, resulting in four rules added to $\pi$.
  The first two steps of regression and lifting are further illustrated in \Cref{fig:moose}.
\end{example}

\subsection{Satisficing Planning with \moose{}}\label{ssec:satisficing}
A learned \moose{} program can be used for satisficing planning by repeatedly choosing and executing a rule to progress the initial state to a goal state.
\Cref{alg:plan} summarises the execution procedure for an input planning problem and \moose{} program.
Each iteration of the algorithm's main loop queries the set of rules in order of ascending precedence values until a rule associated with goals not yet achieved can be grounded (\Crefrange{line:plan:query1}{line:plan:query2}),
from which the corresponding macro action is added to the incumbent plan and applied to the current state (\Crefrange{line:plan:exec1}{line:plan:exec2}).
The loop breaks once the goal is reached (\Cref{line:plan:main}), no actions can be queried from the set of rules, or a cycle is encountered (\Cref{line:plan:failure}).

\newcommand{\ug}{\mathit{ug}}
\newcommand{\x}{\vec{x}}
\newcommand{\y}{\vec{y}}
\subsection{Optimal Planning with \moose{}}\label{ssec:optimal}
Optimal planning can be performed via a synthesised \moose{} program by extending the corresponding planning problem with \moose{} rules.
The rules, ignoring precedence values, are encoded into PDDL axioms~\cite{thiebaux.etal.2005} representing search control for optimal planners that support axioms.
\Cref{thm:opt} later formalises conditions under which encodings of \moose{} programs preserve optimal solutions.

We now extend a given GP domain $\domain=\gen{\predicates, \constants, \schemata}$ with a \moose{} program $\pi$.
We add predicates $p_g$ and $p_{\ug}$ for each $p \in \predicates$, representing goals in a planning problem and unachieved goals in the current state, respectively.
Then each state in a problem $\problem$ is extended with atoms $p_g(\vec{o})$ for each goal atom $p(\vec{o}) \in \problem[g]$ following~\citet{martin.geffner.2004}.
For each predicate $p$ we introduce the axiom
\begin{align*}
  p_{\ug}(\x) \la p_g(\x) \wedge \neg p(\x)
\end{align*}
for computing unachieved goals.
Then for each action schema $a \in \schemata$ we add a new derived predicate $a_\pi$ to $\predicates$ and $\pre(a)$.
Then for each \moose{} rule $\gen{\x, \stateBody, \goalBody, \plan}$, we introduce an axiom
\begin{align*}
  (\plan_1)_\pi(\x) \la \bigwedge_{p(\y) \in \stateBody} p(\y) \wedge \bigwedge_{p(\y) \in \goalBody} p_\ug(\y)
\end{align*}
for restricting the application of an action with the \moose{} rule condition.
In this way, the axioms restrict the set of applicable actions at any ground state to the first action of each macro action that the \moose{} rules would generate, and hence prune the entire search space.
We note that the axioms do not exhibit recursion and thus can be encoded via disjunctive preconditions~\cite{davidson.garagnani.2002}.
Differently to previous works that prune the search space with lifted rules~\cite{bacchus.kabanza.2000,yoon.etal.2008,krajnansky.etal.2014}, our approach does not require writing new solvers but instead makes use of existing planners that support more expressive PDDL features.

\section{Soundness and Completeness Conditions}\label{sec:theory}
In this section, we provide theoretical results concerning the soundness and completeness of \moose{} programs for both satisficing and optimal planning (\Cref{thm:sat,thm:opt}).
Proofs of all statements are provided in \Cref{app:sec:proofs}.
The idea is that under assumptions on the complexity of a GP problem $\gptuple$ and given sufficient training problems, \Cref{alg:learn} synthesises generalised plans from $\trainProblems$ that are sound and complete for solving problems in $\testProblems$, and furthermore finds optimal plans when \moose{} rules are used for search as described in \Cref{ssec:optimal}.
We begin classifying planning domains based on the separability of goals.

\begin{table}
  \newcommand{\complete}{cmpl.}
  \small
  \begin{tabularx}{\columnwidth}{l X X}
    \toprule
    & General Case & \PTIME{} Subplans \\
    \midrule
    TGI \hfill(\ref{defn:tgi}) & \PSPACE{}-\complete \hfill(\ref{thm:tgi}) & in \P{} \hfill(\ref{thm:ptgi}) \\
    SGI \hfill(\ref{defn:sgi}) & \PSPACE{}-\complete \hfill(\ref{thm:sgi}) & \NP{}-\complete \hfill(\ref{thm:psgi}) \\
    OGI \hfill(\ref{defn:ogi}) & \PSPACE{}-\complete \hfill(\ref{thm:ogi}) & \NP{}-\complete \hfill(\ref{thm:pogi}) \\
    \bottomrule
  \end{tabularx}
  \caption{Computational complexity of planning problems exhibiting notions of goal independence. Definitions and theorem references are indicated in brackets.}
  \label{tab:complexity}
\end{table}

\paragraph{Goal Independence}
Early works in planning worked under the assumption that conjunctive goals can be split apart into their individual components and achieved independently.
The \citet{sussman.1973} anomaly illustrates a simple Blocksworld example for how this was not true in general, giving rise to algorithms which aim to achieve goals simultaneously~\cite{waldinger.1977} and to provably complete algorithms in the current planning age.
Regardless, as we see later in our experiments, a non-trivial portion of benchmark planning domains are \P{}-time solvable and furthermore exhibit goals that can be achieved independently from one another.
In this section, we formalise three variants of goal independence (\tgi{}, \sgi{}, \ogi{}) depending on whether serialisation is required and the effects of goal independence on plan optimality.
We further prove their computational complexities summarised in \Cref{tab:complexity}.

Our first notion of goal independence, TGI, describes planning problems that can be solved by solving for each goal conjunct optimally in \emph{any order}.
\begin{definition}[True Goal Independence]\label{defn:tgi}
  A planning problem $\problem$ exhibits true goal independence (\tgi{}) if for all orderings $\vec{g}$ of goal atoms in $\problem[g]$, the following greedy algorithm is sound and complete: (1) set $s = \problem[s_0]$ and then (2) iterate over goal atoms $\vec{g}_i$ in $\vec{g}$ in order by (2a) finding any optimal plan $\plan^{(i)}$ from $s$ to a goal state containing $\vec{g}_i$ and (2b) progressing $s$ via $\plan^{(i)}$.
  We say that $\problem$ exhibits polynomial \tgi{} (\ptgi{}) if step (2a) can run in polynomial time.
  Lastly, we say that $\problem$ exhibits \tgi{} with respect to $C \in \N$, denoted \btgi{}, if all optimal plans in step (2a) have plan length bounded by $C$.
\end{definition}

Our second notion of goal independence, SGI, generalises TGI by relaxing the requirement that the aforementioned greedy algorithm is valid for any order of goal conjuncts. SGI only requires that the greedy algorithm is valid for \emph{at least one order} of goal conjuncts.
\begin{definition}[Serialisable Goal Independence]\label{defn:sgi}
  A planning problem $\problem$ exhibits serialisable goal independence (\sgi{}) if there exists an ordering $\vec{g}$ of goals $\problem[g]$ such that the greedy algorithm operating on $\vec{g}$ is sound and complete.
  Similarly, we say that $\problem$ exhibits polynomial \sgi{} (\psgi{}) if step (2a) in the greedy algorithm runs in polynomial time.
\end{definition}

Our final notion of goal independence, OGI, strengthens the previous independence notion by ensuring that there is at least one order of goal conjuncts for which the greedy algorithm solves the problem \emph{optimally}.
\begin{definition}[Optimal Goal Independence]\label{defn:ogi}
  A planning problem exhibits Optimal Goal Independence (\ogi{}) if there \emph{exists} an ordering $\vec{g}$ of goals $\problem[g]$ such that the algorithm described in \Cref{defn:tgi} is optimally sound and complete with the change that it is now nondeterministic and step (2a) is changed to ``\emph{guess} an optimal plan $\plan^{(i)}$ such that the concatenation of plans is optimal''.
  \pogi{} is defined similarly where (2a) guesses an optimal plan in polynomial time.
\end{definition}

\citet[Sections 4.3 and 4.4]{korf.1987} introduce similar concepts of goal independence corresponding to our TGI and SGI definitions that form the foundation of heuristics for search.
We next theoretically analyse the computational complexity of problems exhibiting the GI definitions with proofs in \Cref{app:sec:proofs}.
We say that a GP problem $\GP = \gptuple$ exhibits \tgi{} if every problem in $\trainProblems \cup \testProblems$ exhibits \tgi{}, and analogously for \sgi{} and \ogi{}.
We then let \plansat{}($\GP$) denote the computational problem of deciding if a plan exists for a problem in $\testProblems$.
The following statements show that without the polynomial time constraint of step (2a) of the aforementioned greedy algorithm, \tgi{} and \sgi{} do not make planning easier.

\begin{proposition}\label{thm:tgi}
  \plansat($\GP$) of a GP problem $\GP$ exhibiting \tgi{} is \PSPACE{}-complete.
\end{proposition}

\begin{corollary}\label{thm:sgi}
  \plansat($\GP$) of a GP problem $\GP$ exhibiting \sgi{} is \PSPACE{}-complete.
\end{corollary}

Once we add the polynomial time constraint of step (2a), both \ptgi{} and \psgi{} become easier.
However, only \ptgi{} becomes tractable while \psgi{} becomes \NP{}-complete.

\begin{proposition}\label{thm:ptgi}
  \plansat($\GP$) of a GP problem $\GP$ exhibiting \ptgi{} is in \P{}.
\end{proposition}

\begin{proposition}\label{thm:psgi}
  \plansat($\GP$) of a GP problem $\GP$ exhibiting \psgi{} is \NP{}-complete.
\end{proposition}

Next, given that SGI is a special case of OGI, we also have the following statements.

\begin{corollary}\label{thm:ogi}
  \plansat($\GP$) of a GP problem $\GP$ exhibiting \ogi{} is \PSPACE{}-complete.
\end{corollary}

\begin{corollary}\label{thm:pogi}
  \plansat($\GP$) of a GP problem $\GP$ exhibiting \pogi{} is \NP{}-complete.
\end{corollary}

\newcommand{\equivalent}{\ensuremath{\sim_U}}

\paragraph{Planning Problem Equivalence}
Before we state the assumptions required for \moose{} to synthesise sound and complete generalised plans, we define the notion of equivalence for (lifted) problems.
We define equivalence via bijection between objects similarly to~\cite{drexler.etal.2024a}, in contrast to work by~\citet{sievers.etal.2019a} which reduces problems to graph automorphisms.
\begin{definition}[Equivalence Relation]\label{defn:equiv}
  Given a GP problem $\gptuple$, we define a relation $\equivalent$ on planning problems in $\trainProblems \cup \testProblems$ by $\problem_1 \equivalent \problem_2$ if there exists a bijective mapping $f: \problem_1[\objects] \to \problem_2[\objects]$ such that $f(c) = c$ for $c \in \constants$, $F(\problem_1[s_0]) = \problem_2[s_0]$ and $F(\problem_1[g]) = \problem_2[g]$ where $F(s) := \set{p(f(o_1), \ldots, f(o_n)) \mid p(o_1, \ldots, o_n) \in s}$.
\end{definition}

Indeed the defined relation is an equivalence relation and furthermore defines a natural notion of equivalence for planning problems, where reflexivity, symmetry and transitivity follows from bijective functions in the definition of $\equivalent$.

\begin{proposition}\label{thm:equivalence}
  The relation $\equivalent$ on planning problems of any given GP problem is an equivalence relation.
\end{proposition}

\begin{proposition}\label{thm:equiv-plan}
  Suppose $\problem_1 \equivalent \problem_2$ and let $f: \problem_1[\objects] \to \problem_2[\objects]$ be the bijective mapping satisfying the definition of $\equivalent$.
  Then a sequence of actions $a_1, \ldots, a_n$ is a plan for $\problem_1$ if and only if $a_1', \ldots, a_n'$ is a plan for $\problem_2$,
  where $a_i'$ is defined by $a_i'=a(f(o_1), \ldots, f(o_n))$ if $a_i=a(o_1, \ldots, o_n)$ for some $a \in \schemata$ and $o_1,\ldots,o_n \in \problem_1[\objects]$.
\end{proposition}

\paragraph{Soundness and Completeness of \moose{}}
Now we state and prove the main theorems of the section.
The main idea of the statement is that given sufficiently many training data, \moose{} can construct a database of rules for \btgi{} problems that can solve all possible problems with singleton goals.
By assuming a bound in the definition of \btgi{} of plan lengths, this database has a finite size which is exponential in the input in the worst case.

\begin{theorem}[\plansat{} soundness and completeness conditions]\label{thm:sat}
  There exists a set $\trainProblems$ such that for all GP problems $\gproblem = \gptuple$ exhibiting \btgi{},
  \Cref{alg:plan} using the plan $\gplan$ synthesised from \Cref{alg:learn} with $\trainProblems$ is sound and complete for $\testProblems$ for satisficing planning.
\end{theorem}

The bound on the size of training problems in the proof of \Cref{thm:sat} is exponential in the domain size.
This bound may be tight and unavoidable given that it has been shown that GP under the QNP~\cite{srivastava.etal.2011} framework is provably equivalent to fully observable non-deterministic (FOND) planning~\cite{bonet.geffner.2020a} which is known to be \EXPTIME{}-complete.
A fruitful next step is to develop a learning algorithm that learns to generate and select what training data is required, possibly given implicitly in the input GP problem~\cite{srivastava.etal.2011a,grundke.etal.2024}.
Now we state the main theorem for optimal planning and provide a counterexample to the theorem for when the OGI assumption is dropped in \cref{eg:ceg} in the Appendix.

\begin{theorem}[\planopt{} soundness and completeness conditions]\label{thm:opt}
  There exists a set $\trainProblems$ such that for all GP problems $\gproblem = \gptuple$ exhibiting \btgi{} and \ogi{},
  an optimal planner run on the transformation in \Cref{ssec:optimal} via the generalised plan $\gplan$ learned from \Cref{alg:learn} with $\trainProblems$ is sound and complete for $\testProblems$ for optimal planning.
\end{theorem}

\newcommand{\qA}{How often can planning benchmark problems be solved by the TGI algorithm (\Cref{defn:tgi})?}
\newcommand{\qB}{How does \moose{} compare to existing generalised planners in \emph{synthesis costs}?}
\newcommand{\qC}{How does \moose{} compare to existing (generalised) planners in \emph{instantiation costs}?}
\newcommand{\qD}{How does \moose{} compare to existing (generalised) planners in \emph{solution quality}?}
\newcommand{\qE}{Can learned \moose{} rules encoded as axioms help speed up existing \emph{optimal planners}?}

\renewcommand{\theenumi}{(\textbf{Q\arabic{enumi}})}
\newif\ifitemize
\itemizefalse

\section{Experiments}\label{sec:experiments}
We conduct experiments to answer the following questions.
\ifitemize
\begin{enumerate}[label=(\textbf{Q\arabic*})]
  \item \label{q:tgi} \qA
  \item \label{q:gp-train} \qB
  \item \label{q:gp-plan} \qC
  \item \label{q:gp-quality} \qD
  \item \label{q:opt} \qE
\end{enumerate}
\else
\refstepcounter{enumi}\label{q:tgi}\theenumi{} \qA{}
\refstepcounter{enumi}\label{q:gp-train}\theenumi{} \qB{}
\refstepcounter{enumi}\label{q:gp-plan}\theenumi{} \qC{}
\refstepcounter{enumi}\label{q:gp-quality}\theenumi{} \qD{}
\refstepcounter{enumi}\label{q:opt}\theenumi{} \qE{}
\fi

\paragraph{\moose{} Implementation}
\moose{} is implemented primarily in Python, but make use of the following tools and planners:
the \texttt{pddl} parser~\cite{favorito.etal.2025} for parsing PDDL problems;
the \texttt{SQLite}~\cite{hipp.2020} database system for grounding in \Cref{alg:plan} as planning states can be viewed as databases and rules as queries~\cite{correa.etal.2020,correa.degiacomo.2024};
\textsc{(Numeric) Fast Downward}'s implementation of A$^*$ with the (Numeric) LM-cut heuristic~\cite{helmert.domshlak.2009,kuroiwa.etal.2022} for generating (numeric) optimal plans in \Cref{line:learn:opt} of \Cref{alg:learn}; and
\symk{}~\cite{speck.etal.2019,speck.etal.2025} for optimal planning with \moose{} rules encoded as axioms described in \Cref{ssec:optimal}.
We further implement an optimisation for generalised plan instantiation that tries to fire the previous successfully fired rule first during \Cref{line:plan:query1,line:plan:query2} of \Cref{alg:plan} to reduce grounding effort.

\paragraph{\ref{q:tgi} \qA}
To answer this question, we use the STRIPS 1998-2023 International Planning Competition (IPC) benchmark suite which consists of 38 domains.
Inspired by the experimental setup by~\citet{lipovetzky.geffner.2012} for testing effective width of planning problems, we test the number of planning problems that can be solved by the TGI algorithm.
For each problem, we randomise a goal ordering and run \textsc{Fast Downward}'s implementation of A$^*$ search with the LM-cut heuristic~\cite{helmert.domshlak.2009} on each goal atom in order as described in \Cref{defn:tgi} with an 8GB memory and 7200s runtime limit.
The output is then validated~\cite{howel.etal.2004}.

In 13 (34.2\%) domains, all returned outputs under the resource constraints are valid, which suggests the possibility that these domains exhibit TGI.
In 19 (50.0\%) domains, at least half of the returned outputs are valid and in 27 (38.0\%) domains, at least one returned output is valid.
In total, of 1184 problems for which an output was returned in the resource constraints, 590 (49.8\%) were valid.
These results suggest that a non-trivial portion of planning domains can be solved by the greedy TGI algorithm.

\paragraph{\ref{q:gp-train} \qB}
To answer this question, we make use of the classical ESHO domains Barman, Ferry, Gripper, Logistics, Miconic, Rovers, Satellite, and Transport, where path-finding components are removed from Rovers and Transport.
Some but not all domains exhibit the \tgi{} assumption.
\Cref{app:sec:benchmarks} provides further benchmark details, distributions of problem object sizes, and number of training and testing problems.
We compare against the sketch learner (\slearner{})~\cite{drexler.etal.2022} with width hyperparameters in $\set{0, 1, 2}$.
All approaches are given a 32GB memory and 12 hour runtime limit for generalised plan synthesis.
Each experiment is repeated 5 times.

\begin{figure*}
  \centering
  \renewcommand{\arraystretch}{0.88}
  \scriptsize
  \tabcolsep 3pt
  \newcommand{\zerocell}{--}
  \newcommand{\www}{0.375\textwidth}
  \begin{tabularx}{\www}{l *{8}{Y}}
  \toprule
  & \multicolumn{4}{c}{Time (s)}
  & \multicolumn{4}{c}{Memory (MB)} \\
  \cmidrule(l){2-5}
  \cmidrule(l){6-9}
  & \header{\slearner-0} & \header{\slearner-1} & \header{\slearner-2} & \header{\moose}
  & \header{\slearner-0} & \header{\slearner-1} & \header{\slearner-2} & \header{\moose} \\
  \cmidrule{1-1}
  \cmidrule(l){2-5}
  \cmidrule(l){6-9}
  Barman & \zerocell & \zerocell & \zerocell & \first{202} & \zerocell & \zerocell & \zerocell & \first{184}\\
  Ferry & 21 & 12 & \first{2} & 9 & 184 & 134 & 76 & \first{52}\\
  Gripper & \first{3} & 9 & 45 & 10 & 66 & 142 & 391 & \first{64}\\
  Logistics & \zerocell & \zerocell & \zerocell & \first{71} & \zerocell & \zerocell & \zerocell & \first{73}\\
  Miconic & 57 & \first{1} & 3 & 12 & 381 & 56 & 125 & \first{52}\\
  Rovers & \zerocell & \zerocell & \zerocell & \first{534} & \zerocell & \zerocell & \zerocell & \first{187}\\
  Satellite & \zerocell & \zerocell & 1559 & \first{514} & \zerocell & \zerocell & 7598 & \first{82}\\
  Transport & \zerocell & \first{12} & \first{12} & 21 & \zerocell & 114 & 129 & \first{80}\\
  \bottomrule
\end{tabularx}

  \hfill
  \newcommand{\ecdfwidth}{0.2\textwidth}
  \raisebox{-0.5\height}{\includegraphics[width=\ecdfwidth]{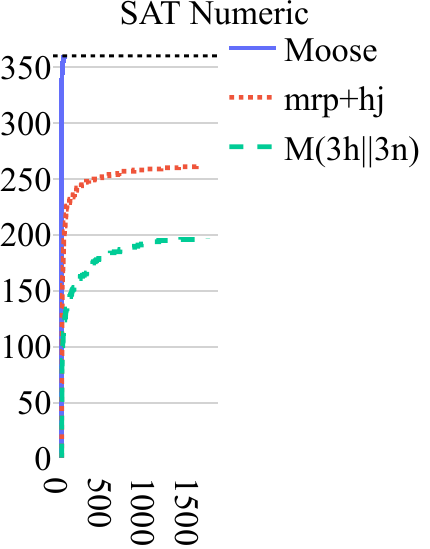}}
  \raisebox{-0.5\height}{\includegraphics[width=\ecdfwidth]{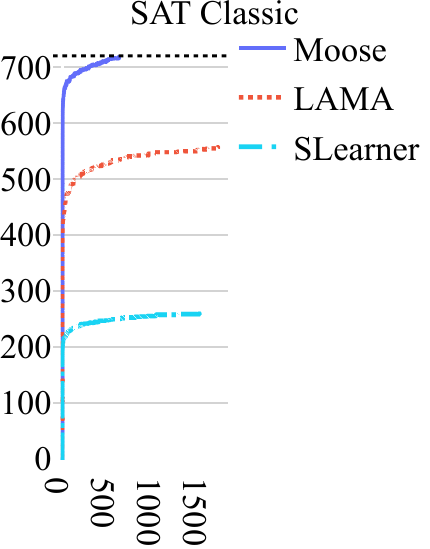}}
  \raisebox{-0.5\height}{\includegraphics[width=\ecdfwidth]{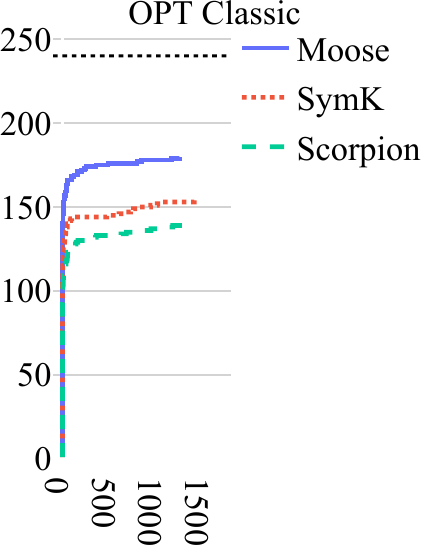}}

  \caption{
    \textbf{Left:} average time and memory usage for synthesis ($\downarrow$). Lowest values are indicated in colour and bold font.
    \textbf{Right:} cumulative coverage ($y$-axis) of planners over time ($x$-axis) for different planning settings. Higher values are better ($\uparrow$).
  }
  \label{tab:synthesis}
  \label{fig:cumulative}
\end{figure*}

\Cref{tab:synthesis} summarises synthesis metrics.
\moose{} completed synthesis in the given synthesis budgets while \slearner{} did not learn domain knowledge for 3 domains with more compute across all hyperparameters.
The best \slearner{} configuration is faster than \moose{} at synthesising generalised plans for simpler domains (Ferry, Miconic and Transport) while \moose{} is faster on 5 out of 8 domains.
However, \moose{} consumes less than 1GB of memory and uses less memory than \slearner{} across all 8 domains.

\paragraph{\ref{q:gp-plan} \qC}
We use the domains and problems from the previous question and the additional numeric planning domains NFerry, NMiconic, NMinecraft, and NTransport.
Further details are again provided in \Cref{app:sec:benchmarks}.
As baselines, we compare against \lama{}~\cite{richter.westphal.2010}, the state-of-the-art standalone satisficing planner in the 2023 IPC Learning Track and the best \slearner{} configuration for every problem for classical planning.
For numeric planning, we compare against the multi-queue (\enhspMQ)~\cite{chen.thiebaux.2024a} and the multi-repetition relaxed plan heuristic with jumping actions (\enhspHMRPHJ)~\cite{scala.etal.2020} configurations of ENHSP.
\moose{} and all baselines are given an 8GB memory and 1800s runtime limit for planning.
\slearner{} and \moose{} experiments are repeated at most 5 times corresponding to the repeated 5 trained models from the previous question.

\Cref{fig:cumulative} displays the cumulative coverage of all planners, where a point $(x, y)$ denotes the number $y$ of problems that can be individually solved in $x$ seconds by a planner.
The dotted black line denotes the number of problems in each benchmark suite.
All 5 seeds of \moose{} solve all numeric planning and classical testing problems except 2 Logistics seeds that fail on a single problem.
In comparison from looking at \Cref{fig:cumulative}, other baseline planners solve $>20$\% fewer problems.
From \Cref{tab:coverage} in \Cref{app:ssec:coverage}, only \moose{} seeds can solve every problem for each domain.

\paragraph{\ref{q:gp-quality} \qD}
We employ the experiments from the previous question and compare \moose{} to the best performing classical and numeric planners \lama{} and \enhspHMRPHJ{}, respectively.
Comparisons are illustrated in \Cref{fig:versus-lama,fig:versus-enhsp} in \Cref{app:ssec:versus-lama,app:ssec:versus-enhsp}.
\moose{} achieves higher quality plans on 3 domains than \enhspHMRPHJ{} (NMiconic, NMinecraft, and NTransport), and worse plan quality on 1 domain (NFerry) for numeric planning.
\moose{} achieves higher quality plans than \lama{} on 5 domains (Barman, Ferry, Miconic, Rovers, Transport), and worse plans on 3 domains (Gripper, Logistics, Satellite) for classical planning.

We further conducted experiments via regressing over goal conjuncts instead of singleton goals in \Cref{alg:learn}.
More rules are synthesised this way leading to higher instantiation costs, but in turn plan quality improved across 7 out of 8 classical domains.
The effects of regressing over goal conjuncts are presented in greater detail in \Cref{app:sec:conjuncts}.

\paragraph{\ref{q:opt} \qE}
For optimal planning, we compare against \scorpion{}~\cite{seipp.etal.2020}, the best standalone optimal planner in the 2023 IPC Optimal Track, and \symk{}~\cite{speck.etal.2025}.
As before, \moose{} and the baselines are given an 8GB memory and 1800s runtime limit.
We note that \scorpion{} and \symk{} are theoretically optimal planners while \moose{} is not guaranteed optimal in practice.

\moose{} solves on average 182.8 out of 240 classical testing problems optimally.
Although the transformation from policies learned from finite training data does not guarantee the preservation of optimal plans, the plans output by \moose{} are empirically optimal.
Both \moose{} and Scorpion achieve the best (tied) performance on 4 domains out of 8.
We also note that \moose{} matches or outperforms the base planner \symk{} on all domains except Gripper.
This observation suggests that the reduction in search space from encoding learned policies via axioms usually outweigh the cost of evaluating such axioms.

\section{Related Work, Discussion and Conclusion}
\citet{gretton.thiebaux.2004} employed regression rewriting for GP in the context of lifted MDPs.
Lifted regression was used to generate relevant features for building decision-tree policies via inductive logic programming.
Their approach handles optimal, probabilistic planning which in turn means that the horizons of testing problems are often bounded by those seen in the training set.
\loom{}~\cite{illanes.mcilraith.2019} automatically synthesises an abstraction from a single planning problem of a GP problem via bagging equivalent objects~\cite{fuentetaja.rosa.2016,riddle.etal.2016,dong.etal.2025} into a nondeterministic, quantified problem.
The quantified problem is then solved with an extension of the FOND planner \prp{}~\cite{muise.etal.2012} to synthesise generalisable policies via regression which satisfy a policy termination test~\cite{srivastava.etal.2011}.
\moose{} takes inspiration from the powerful regression rewriting technique employed in these works, but differs in the methodology.
\moose{} takes a \emph{bottom-up} approach of performing ground regression from example plans to generate ground condition-action pairs that are then lifted into rules.
\citeauthor{gretton.thiebaux.2004} take a \emph{top-down} approach of performing lifted regression to generate relevant lifted features, and \loom{} employs ground regression on a top-down abstraction for synthesising generalised policies.

More generally, \moose{} lies in the class of generalised planners that synthesise generalised plans by sampling from training problems (cf.\ the survey by \citet{celorrio.etal.2019} for more examples).
PG3~\cite{yang.etal.2022}, which performs heuristic search over a space of generalised policies~\cite{segoviaaguas.etal.2021,segoviaaguas.etal.2024}, also uses goal regression for handling `missed' states.
Generalised plans synthesised from sampling have been represented as (deep) policies~\cite{toyer.etal.2018,toyer.etal.2020,bonet.etal.2019,frances.etal.2021,staahlberg.etal.2022,chen.etal.2025}, heuristic functions~\cite{shen.etal.2020,karia.srivastava.2021,chen.etal.2024a,correa.etal.2025} and Python code~\cite{silver.etal.2024}.

In conclusion, we have introduced a new generalised planner, \moose{}, for both satisficing and optimal planning by leveraging goal regression and goal independence.
We formally classify and define the classes of planning domains and problems for which \moose{} is sound and complete.
Experimental results show that our approach significantly advances the state of the art for classical, numeric, and optimal (generalised) planning.
Future work involves extending \moose{} to handle domains requiring transitive closure computations and weaker planning domain assumptions.

\small
\section*{Acknowledgements}
DC and TH carried out this work at the Vector Institute, where DC was a Research Intern.
We gratefully acknowledge funding from the Natural Sciences and Engineering Research Council of Canada (NSERC) and the Canada CIFAR AI Chairs Program.
TH was funded by the Federal Ministry of Education and Research (BMBF) and the Ministry of Culture and Science of the German State of North Rhine-Westphalia (MKW) under the Excellence Strategy of the Federal Government and the Länder.
Resources used in preparing this research were provided, in part, by the Province of Ontario, the Government of Canada through CIFAR, and companies sponsoring the Vector Institute.

\bibliography{support/moose}

@article{aguas.etal.2018,
  _note            = {\gp},
  author           = {Javier Segovia Aguas and Sergio Jim{\'{e}}nez and Anders
                      Jonsson},
  journal          = {J. Artif. Intell. Res.},
  modificationdate = {2025-03-26T14:40:27},
  pages            = {755--797},
  title            = {Computing Hierarchical Finite State Controllers With
                      Classical Planning},
  volume           = {62},
  year             = {2018},
}

@article{bacchus.kabanza.2000,
  _number          = {1-2},
  author           = {Fahiem Bacchus and Froduald Kabanza},
  journal          = {Artif. Intell.},
  modificationdate = {2025-03-26T14:41:09},
  pages            = {123--191},
  title            = {Using temporal logics to express search control knowledge
                      for planning},
  volume           = {116},
  year             = {2000},
}

@article{bonet.geffner.2001,
  author           = {Blai Bonet and Hector Geffner},
  creationdate     = {2024-10-21T10:07:02},
  journal          = {Artif. Intell.},
  modificationdate = {2025-03-26T14:41:09},
  pages            = {5--33},
  title            = {Planning as heuristic search},
  volume           = {129},
  year             = {2001},
}

@article{bonet.geffner.2020a,
  _note            = {\gp},
  author           = {Blai Bonet and Hector Geffner},
  journal          = {J. Artif. Intell. Res.},
  modificationdate = {2025-03-26T14:40:27},
  pages            = {923--961},
  title            = {Qualitative Numeric Planning: Reductions and Complexity},
  volume           = {69},
  year             = {2020},
}

@article{bylander.1994,
  _doi             = {10.1016/0004-3702(94)90081-7},
  _url             = {https://doi.org/10.1016/0004-3702(94)90081-7},
  author           = {Tom Bylander},
  bibsource        = {dblp computer science bibliography, https://dblp.org},
  biburl           = {https://dblp.org/rec/journals/ai/Bylander94.bib},
  journal          = {Artif. Intell.},
  modificationdate = {2025-03-29T10:01:07},
  number           = {1-2},
  pages            = {165--204},
  timestamp        = {Sat, 27 May 2017 14:24:41 +0200},
  title            = {The Computational Complexity of Propositional {STRIPS}
                      Planning},
  volume           = {69},
  year             = {1994},
}

@article{celorrio.etal.2019,
  _doi             = {10.1017/S0269888918000231},
  _note            = {\gp, \survey, \key},
  _url             = {https://doi.org/10.1017/S0269888918000231},
  author           = {Sergio Jim{\'{e}}nez Celorrio and Javier Segovia{-}Aguas
                      and Anders Jonsson},
  bibsource        = {dblp computer science bibliography, https://dblp.org},
  biburl           = {https://dblp.org/rec/journals/ker/CelorrioAJ19.bib},
  journal          = {Knowl. Eng. Rev.},
  modificationdate = {2025-03-26T14:40:27},
  pages            = {e5},
  timestamp        = {Mon, 22 Jul 2024 20:56:21 +0200},
  title            = {A review of generalized planning},
  volume           = {34},
  year             = {2019},
}

@inproceedings{correa.etal.2025,
  author    = {Augusto B. Corrêa and André G. Pereira and Jendrik
               Seipp},
  booktitle = {NeurIPS},
  title     = {Classical Planning with LLM-Generated Heuristics:
               Challenging the State of the Art with Python Code},
  year      = {2025},
}

@article{dong.etal.2025,
  _doi             = {10.1609/aaai.v39i14.33631},
  _url             = {https://ojs.aaai.org/index.php/AAAI/article/view/33631},
  abstractnote     = {Generalized planning is concerned with how to find a
                      single plan to solve multiple similar planning problems.
                      Abstractions are widely used for solving generalized
                      planning, and QNP (qualitative numeric planning) is a
                      popular abstract model. Recently, Cui et al. showed that a
                      plan solves a sound and complete abstraction of a
                      generalized planning problem if and only if the refined
                      plan solves the original problem. However, existing work on
                      automatic abstraction for generalized planning can hardly
                      guarantee soundness let alone completeness. In this paper,
                      we propose an automatic sound and complete abstraction
                      method for generalized planning with baggable types. We use
                      a variant of QNP, called bounded QNP (BQNP), where integer
                      variables are increased or decreased by only one. Since
                      BQNP is undecidable, we propose and implement a sound but
                      incomplete solver for BQNP. We present an automatic method
                      to abstract a BQNP problem from a classical planning
                      problem with baggable types. The basic idea for abstraction
                      is to introduce a counter for each bag of indistinguishable
                      tuples of objects. We define a class of domains called
                      proper baggable domains, and show that for such domains,
                      the BQNP problem got by our automatic method is a sound and
                      complete abstraction for a generalized planning problem
                      whose instances share the same bags with the given instance
                      but the sizes of the bags might be different. Thus, the
                      refined plan of a solution to the BQNP problem is a
                      solution to the generalized planning problem. Finally, we
                      implement our abstraction method and experiments on a
                      number of domains demonstrate the promise of our
                      approach.},
  author           = {Dong, Hao and Shi, Zheyuan and Zeng, Hemeng and Liu,
                      Yongmei},
  journal          = {AAAI},
  modificationdate = {2025-04-28T10:12:26},
  number           = {14},
  pages            = {14875-14884},
  title            = {An Automatic Sound and Complete Abstraction Method for
                      Generalized Planning with Baggable Types},
  volume           = {39},
  year             = {2025},
}

@article{erol.etal.1995,
  _number          = {1-2},
  author           = {Kutluhan Erol and Dana S. Nau and V. S. Subrahmanian},
  creationdate     = {2024-10-30T10:17:52},
  journal          = {Artif. Intell.},
  modificationdate = {2025-03-26T14:41:09},
  pages            = {75--88},
  title            = {Complexity, Decidability and Undecidability Results for
                      Domain-Independent Planning},
  volume           = {76},
  year             = {1995},
}

@article{fikes.etal.1972,
  _doi             = {10.1016/0004-3702(72)90051-3},
  _url             = {https://doi.org/10.1016/0004-3702(72)90051-3},
  author           = {Richard Fikes and Peter E. Hart and Nils J. Nilsson},
  bibsource        = {dblp computer science bibliography, https://dblp.org},
  biburl           = {https://dblp.org/rec/journals/ai/FikesHN72.bib},
  journal          = {Artif. Intell.},
  modificationdate = {2025-05-01T10:14:56},
  number           = {1-3},
  pages            = {251--288},
  timestamp        = {Sat, 27 May 2017 14:24:42 +0200},
  title            = {Learning and Executing Generalized Robot Plans},
  volume           = {3},
  year             = {1972},
}

@article{fox.long.2003,
  _pages           = {61-124},
  author           = {Maria Fox and Derek Long},
  creationdate     = {2024-11-06T13:54:26},
  journal          = {J. Artif. Intell. Res.},
  modificationdate = {2025-03-26T14:40:28},
  title            = {{PDDL2.1:} An Extension to {PDDL} for Expressing Temporal
                      Planning Domains},
  volume           = {20},
  year             = {2003},
}

@article{fuentetaja.rosa.2016,
  _doi             = {10.3233/AIC-150692},
  _url             = {https://doi.org/10.3233/AIC-150692},
  author           = {Raquel Fuentetaja and Tom{\'{a}}s de la Rosa},
  bibsource        = {dblp computer science bibliography, https://dblp.org},
  biburl           = {https://dblp.org/rec/journals/aicom/FuentetajaR15.bib},
  journal          = {{AI} Commun.},
  modificationdate = {2025-05-18T13:36:30},
  number           = {3},
  pages            = {435--467},
  timestamp        = {Thu, 25 Nov 2021 09:28:47 +0100},
  title            = {Compiling irrelevant objects to counters. Special case of
                      creation planning},
  volume           = {29},
  year             = {2016},
}

@article{hearn.demaine.2005,
  _doi             = {10.1016/J.TCS.2005.05.008},
  _url             = {https://doi.org/10.1016/j.tcs.2005.05.008},
  author           = {Robert A. Hearn and Erik D. Demaine},
  bibsource        = {dblp computer science bibliography, https://dblp.org},
  biburl           = {https://dblp.org/rec/journals/tcs/HearnD05.bib},
  journal          = {Theor. Comput. Sci.},
  modificationdate = {2025-03-29T10:40:19},
  number           = {1-2},
  pages            = {72--96},
  timestamp        = {Wed, 17 Feb 2021 22:00:03 +0100},
  title            = {PSPACE-completeness of sliding-block puzzles and other
                      problems through the nondeterministic constraint logic
                      model of computation},
  volume           = {343},
  year             = {2005},
}

@article{helmert.2003,
  _doi             = {10.1016/S0004-3702(02)00364-8},
  _url             = {https://doi.org/10.1016/S0004-3702(02)00364-8},
  author           = {Malte Helmert},
  bibsource        = {dblp computer science bibliography, https://dblp.org},
  biburl           = {https://dblp.org/rec/journals/ai/Helmert03.bib},
  journal          = {Artif. Intell.},
  modificationdate = {2025-05-08T11:42:15},
  number           = {2},
  pages            = {219--262},
  timestamp        = {Sun, 06 Oct 2024 21:18:16 +0200},
  title            = {Complexity results for standard benchmark domains in
                      planning},
  volume           = {143},
  year             = {2003},
}

@article{khardon.1999,
  _note            = {\lp},
  _number          = {1-2},
  author           = {Roni Khardon},
  journal          = {Artif. Intell.},
  modificationdate = {2025-03-26T14:40:27},
  pages            = {125--148},
  title            = {Learning Action Strategies for Planning Domains},
  volume           = {113},
  year             = {1999},
}

@article{kuroiwa.etal.2022,
  _doi             = {10.1613/JAIR.1.14034},
  _url             = {https://doi.org/10.1613/jair.1.14034},
  author           = {Ryo Kuroiwa and Alexander Shleyfman and Chiara Piacentini
                      and Margarita P. Castro and J. Christopher Beck},
  bibsource        = {dblp computer science bibliography, https://dblp.org},
  biburl           = {https://dblp.org/rec/journals/jair/KuroiwaSPCB22.bib},
  journal          = {J. Artif. Intell. Res.},
  modificationdate = {2025-04-16T11:22:52},
  pages            = {1477--1548},
  timestamp        = {Sun, 12 Nov 2023 02:18:28 +0100},
  title            = {The {LM}-Cut Heuristic Family for Optimal Numeric Planning
                      with Simple Conditions},
  volume           = {75},
  year             = {2022},
}

@article{martin.geffner.2004,
  _note            = {\gp},
  _number          = {1},
  author           = {Mario Mart{\'{\i}}n and Hector Geffner},
  creationdate     = {2024-10-21T08:04:27},
  journal          = {Appl. Intell.},
  modificationdate = {2025-03-26T14:40:27},
  pages            = {9--19},
  title            = {Learning Generalized Policies from Planning Examples Using
                      Concept Languages},
  volume           = {20},
  year             = {2004},
}

@article{richter.westphal.2010,
  author           = {Silvia Richter and Matthias Westphal},
  creationdate     = {2024-10-30T12:39:06},
  journal          = {J. Artif. Intell. Res.},
  modificationdate = {2025-03-26T14:41:09},
  pages            = {127--177},
  title            = {The {LAMA} Planner: Guiding Cost-Based Anytime Planning
                      with Landmarks},
  volume           = {39},
  year             = {2010},
}

@article{sanner.boutilier.2009,
  _doi             = {10.1016/J.ARTINT.2008.11.003},
  _url             = {https://doi.org/10.1016/j.artint.2008.11.003},
  author           = {Scott Sanner and Craig Boutilier},
  bibsource        = {dblp computer science bibliography, https://dblp.org},
  biburl           = {https://dblp.org/rec/journals/ai/SannerB09.bib},
  journal          = {Artif. Intell.},
  modificationdate = {2025-05-29T10:35:54},
  number           = {5-6},
  pages            = {748--788},
  timestamp        = {Sat, 27 May 2017 14:24:42 +0200},
  title            = {Practical solution techniques for first-order {MDP}s},
  volume           = {173},
  year             = {2009},
}

@article{segoviaaguas.etal.2024,
  _note            = {\gp},
  author           = {Javier Segovia{-}Aguas and Sergio Jim{\'{e}}nez Celorrio
                      and Anders Jonsson},
  creationdate     = {2024-09-23T10:52:22},
  journal          = {Artif. Intell.},
  modificationdate = {2025-03-26T14:40:27},
  pages            = {104097},
  title            = {Generalized planning as heuristic search: {A} new planning
                      search-space that leverages pointers over objects},
  volume           = {330},
  year             = {2024},
}

@article{seipp.etal.2020,
  author           = {Jendrik Seipp and Thomas Keller and Malte Helmert},
  journal          = {J. Artif. Intell. Res.},
  modificationdate = {2025-03-26T14:41:09},
  pages            = {129--167},
  title            = {Saturated Cost Partitioning for Optimal Classical
                      Planning},
  volume           = {67},
  year             = {2020},
}

@article{speck.etal.2025,
  _doi             = {10.1613/JAIR.1.13599},
  _note            = {\gp},
  _url             = {https://doi.org/10.1613/jair.1.13599},
  author           = {David Speck and Jendrik Seipp and {\'{A}}lvaro Torralba},
  bibsource        = {dblp computer science bibliography, https://dblp.org},
  biburl           = {https://dblp.org/rec/journals/jair/RodriguezBSG22.bib},
  journal          = {J. Artif. Intell. Res.},
  modificationdate = {2025-03-28T11:04:41},
  pages            = {1349--1405},
  timestamp        = {Mon, 28 Aug 2023 21:18:42 +0200},
  title            = {Symbolic Search for Cost-Optimal Planning with Expressive
                      Model Extensions},
  volume           = {82},
  year             = {2025},
}

@article{srivastava.etal.2011a,
  _doi             = {10.1016/J.ARTINT.2010.10.006},
  _note            = {\gp},
  _url             = {https://doi.org/10.1016/j.artint.2010.10.006},
  author           = {Siddharth Srivastava and Neil Immerman and Shlomo
                      Zilberstein},
  bibsource        = {dblp computer science bibliography, https://dblp.org},
  biburl           = {https://dblp.org/rec/journals/ai/SrivastavaIZ11.bib},
  journal          = {Artif. Intell.},
  modificationdate = {2025-03-26T14:40:27},
  number           = {2},
  pages            = {615--647},
  timestamp        = {Sun, 05 May 2024 12:43:27 +0200},
  title            = {A new representation and associated algorithms for
                      generalized planning},
  volume           = {175},
  year             = {2011},
}

@article{taitler.etal.2024,
  _number          = {2},
  author           = {Ayal Taitler and Ron Alford and Joan Espasa and Gregor
                      Behnke and Daniel Fiser and Michael Gimelfarb and Florian
                      Pommerening and Scott Sanner and Enrico Scala and Dominik
                      Schreiber and Javier Segovia{-}Aguas and Jendrik Seipp},
  creationdate     = {2024-09-06T20:02:54},
  journal          = {{AI} Mag.},
  modificationdate = {2025-03-26T14:41:09},
  pages            = {280--296},
  title            = {The 2023 International Planning Competition},
  volume           = {45},
  year             = {2024},
}

@article{thiebaux.etal.2005,
  _number          = {1-2},
  author           = {Sylvie Thi{\'{e}}baux and J{\"{o}}rg Hoffmann and Bernhard
                      Nebel},
  journal          = {Artif. Intell.},
  modificationdate = {2025-03-26T14:41:09},
  pages            = {38--69},
  title            = {In defense of {PDDL} axioms},
  volume           = {168},
  year             = {2005},
}

@article{toyer.etal.2020,
  _note            = {\lp},
  author           = {Sam Toyer and Sylvie Thi{\'{e}}baux and Felipe Trevizan
                      and Lexing Xie},
  journal          = {J. Artif. Intell. Res.},
  modificationdate = {2025-03-26T14:40:27},
  pages            = {1--68},
  title            = {ASNets: Deep Learning for Generalised Planning},
  volume           = {68},
  year             = {2020},
}

@article{waldinger.1977,
  author           = {Richard J. Waldinger},
  journal          = {Machine Intelligence},
  modificationdate = {2025-03-29T09:59:36},
  pages            = {94--136},
  school           = {MIT},
  title            = {Achieving several goals simultaneously},
  volume           = {8},
  year             = {1977},
}

@article{yoon.etal.2008,
  _note            = {\lp},
  author           = {Sung Wook Yoon and Alan Fern and Robert Givan},
  creationdate     = {2024-09-23T09:46:55},
  journal          = {J. Mach. Learn. Res.},
  modificationdate = {2025-03-26T14:40:27},
  pages            = {683--718},
  title            = {Learning Control Knowledge for Forward Search Planning},
  volume           = {9},
  year             = {2008},
}

@book{garey.johnson.1979,
  author           = {M. R. Garey and David S. Johnson},
  bibsource        = {dblp computer science bibliography, https://dblp.org},
  isbn             = {0-7167-1044-7},
  modificationdate = {2025-03-29T11:00:52},
  publisher        = {W. H. Freeman},
  title            = {Computers and Intractability: {A} Guide to the Theory of
                      NP-Completeness},
  year             = {1979},
}

@book{haslum.etal.2019,
  _series          = {Synthesis Lectures on Artificial Intelligence and Machine
                      Learning},
  author           = {Patrik Haslum and Nir Lipovetzky and Daniele Magazzeni and
                      Christian Muise},
  creationdate     = {2024-10-04T12:36:50},
  modificationdate = {2025-03-26T14:41:09},
  publisher        = {Morgan {\&} Claypool Publishers},
  title            = {An Introduction to the Planning Domain Definition
                      Language},
  year             = {2019},
}

@book{reiter.2001,
  author           = {Raymond Reiter},
  modificationdate = {2025-03-30T08:43:00},
  publisher        = {MIT Press},
  title            = {Knowledge in Action: Logical Foundations for Specifying
                      and Implementing Dynamical Systems},
  year             = {2001},
}

@incollection{reiter.1991,
  _doi             = {https://doi.org/10.1016/B978-0-12-450010-5.50026-8},
  _url             = {https://www.sciencedirect.com/science/article/pii/B9780124500105500268},
  author           = {Raymond Reiter},
  booktitle        = {Artificial and Mathematical Theory of Computation},
  modificationdate = {2025-05-18T14:01:37},
  publisher        = {Academic Press},
  title            = {The Frame Problem in the Situation Calculus: A Simple
                      Solution (Sometimes) and a Completeness Result for Goal
                      Regression},
  year             = {1991},
}

@inproceedings{alcazar.etal.2013,
  _editor          = {Francesca Rossi},
  _pages           = {2254--2260},
  _publisher       = {{IJCAI/AAAI}},
  _url             = {http://www.aaai.org/ocs/index.php/IJCAI/IJCAI13/paper/view/6877},
  author           = {Vidal Alc{\'{a}}zar and Daniel Borrajo and Susana
                      Fern{\'{a}}ndez and Raquel Fuentetaja},
  bibsource        = {dblp computer science bibliography, https://dblp.org},
  biburl           = {https://dblp.org/rec/conf/ijcai/AlcazarBFF13.bib},
  booktitle        = {IJCAI},
  modificationdate = {2025-03-28T16:08:06},
  timestamp        = {Tue, 23 Jan 2024 13:25:46 +0100},
  title            = {Revisiting Regression in Planning},
  year             = {2013},
}

@inproceedings{benyamin.etal.2024,
  author           = {Benyamin, Yarin and Mordoch, Argaman and Shperberg, Shahaf
                      and Stern, Roni},
  booktitle        = {Proceedings of the PRL Workshop Series: Bridging the Gap Between AI Planning
                      and Reinforcement Learning},
  modificationdate = {2025-04-16T13:46:30},
  title            = {Solving Minecraft Tasks via Model Learning},
  year             = {2024},
}

@inproceedings{bonet.etal.2009,
  _note            = {\gp},
  _pages           = {34--41},
  author           = {Blai Bonet and H{\'{e}}ctor Palacios and Hector Geffner},
  booktitle        = {{ICAPS}},
  modificationdate = {2025-03-26T14:40:27},
  title            = {Automatic Derivation of Memoryless Policies and
                      Finite-State Controllers Using Classical Planners},
  year             = {2009},
}

@inproceedings{bonet.etal.2010,
  _note            = {\gp},
  _pages           = {1656--1659},
  author           = {Blai Bonet and H{\'{e}}ctor Palacios and Hector Geffner},
  booktitle        = {{AAAI}},
  modificationdate = {2025-03-26T14:40:27},
  title            = {Automatic Derivation of Finite-State Machines for Behavior
                      Control},
  year             = {2010},
}

@inproceedings{bonet.etal.2019,
  _note            = {\gp},
  _pages           = {2703--2710},
  author           = {Blai Bonet and Guillem Franc{\`{e}}s and Hector Geffner},
  booktitle        = {{AAAI}},
  creationdate     = {2024-09-24T17:19:50},
  modificationdate = {2025-03-26T14:40:27},
  title            = {Learning Features and Abstract Actions for Computing
                      Generalized Plans},
  year             = {2019},
}

@inproceedings{chen.etal.2024a,
  _note            = {\lp},
  _pages           = {20078--20086},
  author           = {Dillon Z. Chen and Sylvie Thi{\'{e}}baux and Felipe
                      Trevizan},
  booktitle        = {AAAI},
  modificationdate = {2025-03-26T14:40:27},
  title            = {Learning Domain-Independent Heuristics for Grounded and
                      Lifted Planning},
  year             = {2024},
}

@inproceedings{chen.thiebaux.2024,
  _note            = {\lp},
  author           = {Dillon Z. Chen and Sylvie Thi{\'{e}}baux},
  booktitle        = {NeurIPS},
  creationdate     = {2024-10-21T08:24:15},
  modificationdate = {2025-03-26T14:40:27},
  title            = {Graph Learning for Numeric Planning},
  year             = {2024},
}

@inproceedings{chen.etal.2025,
  author    = {Dillon Z. Chen and Johannes Zenn and Tristan Cinquin and Sheila A. McIlraith},
  booktitle = {Proceedings of the 18th European Workshop on Reinforcement Learning (EWRL)},
  title     = {Language Models For Generalised PDDL Planning: Synthesising Sound and Programmatic Policies},
  year      = {2025},
}

@inproceedings{chen.thiebaux.2024a,
  _doi             = {10.1609/SOCS.V17I1.31559},
  _editor          = {Ariel Felner and Jiaoyang Li},
  _pages           = {203--207},
  _publisher       = {{AAAI} Press},
  _url             = {https://doi.org/10.1609/socs.v17i1.31559},
  author           = {Dillon Z. Chen and Sylvie Thi{\'{e}}baux},
  bibsource        = {dblp computer science bibliography, https://dblp.org},
  biburl           = {https://dblp.org/rec/conf/socs/ChenT24.bib},
  booktitle        = {SOCS},
  modificationdate = {2025-04-16T13:50:24},
  timestamp        = {Mon, 03 Jun 2024 17:01:43 +0200},
  title            = {Novelty Heuristics, Multi-Queue Search, and Portfolios for
                      Numeric Planning},
  year             = {2024},
}

@inproceedings{correa.degiacomo.2024,
  _pages           = {8010--8019},
  _publisher       = {ijcai.org},
  _url             = {https://www.ijcai.org/proceedings/2024/886},
  author           = {Augusto B. Corr{\^{e}}a and Giuseppe {De Giacomo}},
  bibsource        = {dblp computer science bibliography, https://dblp.org},
  biburl           = {https://dblp.org/rec/conf/ijcai/CorreaG24.bib},
  booktitle        = {IJCAI},
  modificationdate = {2025-05-22T12:43:46},
  timestamp        = {Fri, 18 Oct 2024 20:55:24 +0200},
  title            = {Lifted Planning: Recent Advances in Planning Using
                      First-Order Representations},
  year             = {2024},
}

@inproceedings{correa.etal.2020,
  _editor          = {J. Christopher Beck and Olivier Buffet and J{\"{o}}rg
                      Hoffmann and Erez Karpas and Shirin Sohrabi},
  _pages           = {80--89},
  _publisher       = {{AAAI} Press},
  _url             = {https://ojs.aaai.org/index.php/ICAPS/article/view/6648},
  author           = {Augusto B. Corr{\^{e}}a and Florian Pommerening and Malte
                      Helmert and Guillem Franc{\`{e}}s},
  bibsource        = {dblp computer science bibliography, https://dblp.org},
  biburl           = {https://dblp.org/rec/conf/aips/CorreaPHF20.bib},
  booktitle        = {ICAPS},
  modificationdate = {2025-05-22T12:44:08},
  timestamp        = {Mon, 07 Mar 2022 16:58:34 +0100},
  title            = {Lifted Successor Generation Using Query Optimization
                      Techniques},
  year             = {2020},
}

@inproceedings{drexler.etal.2022,
  _editor          = {Akshat Kumar and Sylvie Thi{\'{e}}baux and Pradeep
                      Varakantham and William Yeoh},
  _pages           = {62--70},
  _publisher       = {{AAAI} Press},
  _url             = {https://ojs.aaai.org/index.php/ICAPS/article/view/19786},
  author           = {Dominik Drexler and Jendrik Seipp and Hector Geffner},
  bibsource        = {dblp computer science bibliography, https://dblp.org},
  biburl           = {https://dblp.org/rec/conf/aips/DrexlerSG22.bib},
  booktitle        = {ICAPS},
  modificationdate = {2025-04-18T14:11:54},
  timestamp        = {Wed, 20 Jul 2022 14:27:38 +0200},
  title            = {Learning Sketches for Decomposing Planning Problems into
                      Subproblems of Bounded Width},
  year             = {2022},
}

@inproceedings{drexler.etal.2024a,
  author           = {Dominik Drexler and Simon St{\aa}hlberg and Blai Bonet and
                      Hector Geffner},
  booktitle        = {{KR}},
  creationdate     = {2024-08-31T16:12:56},
  modificationdate = {2025-03-26T14:40:27},
  title            = {Equivalence-Based Abstractions for Learning General
                      Policies},
  year             = {2024},
}

@inproceedings{frances.etal.2021,
  _doi             = {10.1609/AAAI.V35I13.17402},
  _pages           = {11801--11808},
  _publisher       = {{AAAI} Press},
  _url             = {https://doi.org/10.1609/aaai.v35i13.17402},
  author           = {Guillem Franc{\`{e}}s and Blai Bonet and Hector Geffner},
  bibsource        = {dblp computer science bibliography, https://dblp.org},
  biburl           = {https://dblp.org/rec/conf/aaai/FrancesBG21.bib},
  booktitle        = {AAAI},
  modificationdate = {2025-03-26T14:40:27},
  timestamp        = {Mon, 04 Sep 2023 16:50:24 +0200},
  title            = {Learning General Planning Policies from Small Examples
                      Without Supervision},
  year             = {2021},
}

@inproceedings{fritz.mcilraith.2007,
  _editor          = {Mark S. Boddy and Maria Fox and Sylvie Thi{\'{e}}baux},
  _pages           = {144--151},
  _publisher       = {{AAAI}},
  _url             = {http://www.aaai.org/Library/ICAPS/2007/icaps07-019.php},
  author           = {Christian Fritz and Sheila A. McIlraith},
  bibsource        = {dblp computer science bibliography, https://dblp.org},
  biburl           = {https://dblp.org/rec/conf/aips/FritzM07.bib},
  booktitle        = {ICAPS},
  modificationdate = {2025-03-28T16:07:59},
  timestamp        = {Tue, 02 Nov 2021 15:59:05 +0100},
  title            = {Monitoring Plan Optimality During Execution},
  year             = {2007},
}

@inproceedings{gretton.thiebaux.2004,
  _note            = {\lp},
  _pages           = {217--225},
  author           = {Charles Gretton and Sylvie Thi{\'{e}}baux},
  booktitle        = {{UAI}},
  creationdate     = {2024-09-13T10:50:53},
  modificationdate = {2025-03-26T14:40:27},
  title            = {Exploiting First-Order Regression in Inductive Policy
                      Selection},
  year             = {2004},
}

@inproceedings{grundke.etal.2024,
  _pages           = {239--248},
  author           = {Claudia Grundke and Gabriele R{\"{o}}ger and Malte
                      Helmert},
  booktitle        = {{ICAPS}},
  modificationdate = {2025-03-26T14:40:27},
  title            = {Formal Representations of Classical Planning Domains},
  year             = {2024},
}

@inproceedings{helmert.2002,
  _editor          = {Malik Ghallab and Joachim Hertzberg and Paolo Traverso},
  _pages           = {44--53},
  _publisher       = {{AAAI}},
  _url             = {http://www.aaai.org/Library/AIPS/2002/aips02-005.php},
  author           = {Malte Helmert},
  bibsource        = {dblp computer science bibliography, https://dblp.org},
  biburl           = {https://dblp.org/rec/conf/aips/Helmert02.bib},
  booktitle        = {AIPS},
  modificationdate = {2025-03-26T14:40:27},
  timestamp        = {Fri, 05 Feb 2021 17:14:50 +0100},
  title            = {Decidability and Undecidability Results for Planning with
                      Numerical State Variables},
  year             = {2002},
}

@inproceedings{helmert.domshlak.2009,
  _editor          = {Alfonso Gerevini and Adele E. Howe and Amedeo Cesta and
                      Ioannis Refanidis},
  _publisher       = {{AAAI}},
  _url             = {http://aaai.org/ocs/index.php/ICAPS/ICAPS09/paper/view/735},
  author           = {Malte Helmert and Carmel Domshlak},
  bibsource        = {dblp computer science bibliography, https://dblp.org},
  biburl           = {https://dblp.org/rec/conf/aips/HelmertD09.bib},
  booktitle        = {ICAPS},
  modificationdate = {2025-03-28T16:07:42},
  timestamp        = {Thu, 13 Dec 2012 14:15:16 +0100},
  title            = {Landmarks, Critical Paths and Abstractions: What's the
                      Difference Anyway?},
  year             = {2009},
}

@inproceedings{hofmann.etal.2020,
  _doi             = {10.3233/FAIA200164},
  _pages           = {761--768},
  _url             = {https://doi.org/10.3233/FAIA200164},
  author           = {Till Hofmann and Tim Niemueller and Gerhard Lakemeyer},
  bibsource        = {dblp computer science bibliography, https://dblp.org},
  biburl           = {https://dblp.org/rec/conf/ecai/HofmannNL20.bib},
  booktitle        = {ECAI},
  modificationdate = {2025-05-08T13:59:39},
  timestamp        = {Fri, 09 Apr 2021 18:50:05 +0200},
  title            = {Macro Operator Synthesis for {ADL} Domains},
  year             = {2020},
}

@inproceedings{hu.giacomo.2011,
  _note            = {\gp},
  _pages           = {918--923},
  author           = {Yuxiao Hu and Giuseppe {De Giacomo}},
  booktitle        = {{IJCAI}},
  modificationdate = {2025-05-05T10:11:13},
  title            = {Generalized Planning: Synthesizing Plans that Work for
                      Multiple Environments},
  year             = {2011},
}

@inproceedings{illanes.mcilraith.2017,
  _pages    = {4338--4345},
  author    = {Leon Illanes and Sheila A. McIlraith},
  booktitle = {{IJCAI}},
  title     = {Numeric Planning via Abstraction and Policy Guided
               Search},
  year      = {2017},
}

@inproceedings{illanes.mcilraith.2019,
  _note            = {\gp},
  _pages           = {7610--7618},
  author           = {Le{\'{o}}n Illanes and Sheila A. McIlraith},
  booktitle        = {{AAAI}},
  modificationdate = {2025-03-26T14:40:27},
  title            = {Generalized Planning via Abstraction: Arbitrary Numbers of
                      Objects},
  year             = {2019},
}

@inproceedings{karia.srivastava.2021,
  _note            = {\lp},
  _pages           = {8064--8073},
  author           = {Rushang Karia and Siddharth Srivastava},
  booktitle        = {AAAI},
  creationdate     = {2024-08-29T17:23:01},
  modificationdate = {2025-03-26T14:40:27},
  title            = {Learning Generalized Relational Heuristic Networks for
                      Model-Agnostic Planning},
  year             = {2021},
}

@inproceedings{krajnansky.etal.2014,
  _doi             = {10.3233/978-1-61499-419-0-483},
  _editor          = {Torsten Schaub and Gerhard Friedrich and Barry
                      O'Sullivan},
  _pages           = {483--488},
  _publisher       = {{IOS} Press},
  _series          = {Frontiers in Artificial Intelligence and Applications},
  _url             = {https://doi.org/10.3233/978-1-61499-419-0-483},
  _volume          = {263},
  author           = {Michal Krajnansk{\'{y}} and J{\"{o}}rg Hoffmann and
                      Olivier Buffet and Alan Fern},
  bibsource        = {dblp computer science bibliography, https://dblp.org},
  biburl           = {https://dblp.org/rec/conf/ecai/Krajnansky0BF14.bib},
  booktitle        = {{ECAI}},
  modificationdate = {2025-01-07T19:45:22},
  timestamp        = {Mon, 19 Jun 2023 16:36:09 +0200},
  title            = {Learning Pruning Rules for Heuristic Search Planning},
  year             = {2014},
}

@inproceedings{levesque.2005,
  _note            = {\gp},
  _pages           = {509--515},
  author           = {Levesque, H. J.},
  booktitle        = {IJCAI},
  modificationdate = {2025-03-26T14:40:27},
  title            = {Planning with Loops},
  year             = {2005},
}

@inproceedings{lipovetzky.geffner.2012,
  author    = {Nir Lipovetzky and Hector Geffner},
  booktitle = {{ECAI}},
  title     = {Width and Serialization of Classical Planning Problems},
  year      = {2012},
}

@inproceedings{muise.etal.2012,
  _editor          = {Lee McCluskey and Brian Charles Williams and Jos{\'{e}}
                      Reinaldo Silva and Blai Bonet},
  _publisher       = {{AAAI}},
  _url             = {http://www.aaai.org/ocs/index.php/ICAPS/ICAPS12/paper/view/4718},
  author           = {Christian J. Muise and Sheila A. McIlraith and J.
                      Christopher Beck},
  bibsource        = {dblp computer science bibliography, https://dblp.org},
  biburl           = {https://dblp.org/rec/conf/aips/MuiseMB12.bib},
  booktitle        = {ICAPS},
  modificationdate = {2025-03-26T14:40:27},
  timestamp        = {Wed, 01 Mar 2017 18:10:41 +0100},
  title            = {Improved Non-Deterministic Planning by Exploiting State
                      Relevance},
  year             = {2012},
}

@inproceedings{muise.etal.2024,
  _doi             = {10.1609/AAAI.V38I18.30001},
  _editor          = {Michael J. Wooldridge and Jennifer G. Dy and Sriraam
                      Natarajan},
  _pages           = {20212--20221},
  _publisher       = {{AAAI} Press},
  _url             = {https://doi.org/10.1609/aaai.v38i18.30001},
  author           = {Christian Muise and Sheila A. McIlraith and J. Christopher
                      Beck},
  bibsource        = {dblp computer science bibliography, https://dblp.org},
  biburl           = {https://dblp.org/rec/conf/aaai/MuiseMB24.bib},
  booktitle        = {AAAI},
  modificationdate = {2025-03-28T16:08:02},
  timestamp        = {Tue, 02 Apr 2024 16:32:09 +0200},
  title            = {{PRP} Rebooted: Advancing the State of the Art in {FOND}
                      Planning},
  year             = {2024},
}

@inproceedings{pang.holte.2011,
  _doi             = {10.1609/SOCS.V2I1.18189},
  _editor          = {Daniel Borrajo and Maxim Likhachev and Carlos Linares
                      L{\'{o}}pez},
  _pages           = {125--133},
  _publisher       = {{AAAI} Press},
  _url             = {https://doi.org/10.1609/socs.v2i1.18189},
  author           = {Bo Pang and Robert C. Holte},
  bibsource        = {dblp computer science bibliography, https://dblp.org},
  biburl           = {https://dblp.org/rec/conf/socs/PangH11.bib},
  booktitle        = {SOCS},
  modificationdate = {2025-03-28T16:08:10},
  timestamp        = {Mon, 18 Dec 2023 16:58:34 +0100},
  title            = {State-Set Search},
  year             = {2011},
}

@inproceedings{riddle.etal.2016,
  author           = {Riddle, Pat and Douglas, Jordan and Barley, Mike and
                      Franco, Santiago},
  booktitle        = {Proceedings of the 8th Workshop on Heuristic Search for
                      Domain-independent Planning},
  modificationdate = {2025-05-18T14:02:06},
  title            = {Improving performance by reformulating {PDDL} into a bagged
                      representation},
  year             = {2016},
}

@inproceedings{scala.etal.2016,
  _editor          = {Subbarao Kambhampati},
  _pages           = {3228--3234},
  _publisher       = {{IJCAI/AAAI} Press},
  _url             = {http://www.ijcai.org/Abstract/16/457},
  author           = {Enrico Scala and Patrik Haslum and Sylvie Thi{\'{e}}baux},
  bibsource        = {dblp computer science bibliography, https://dblp.org},
  biburl           = {https://dblp.org/rec/conf/ijcai/ScalaHT16.bib},
  booktitle        = {IJCAI},
  modificationdate = {2025-04-17T19:28:36},
  timestamp        = {Tue, 20 Aug 2019 16:19:00 +0200},
  title            = {Heuristics for Numeric Planning via Subgoaling},
  year             = {2016},
}

@inproceedings{scala.etal.2020,
  _editor          = {J. Christopher Beck and Olivier Buffet and J{\"{o}}rg
                      Hoffmann and Erez Karpas and Shirin Sohrabi},
  _pages           = {226--234},
  _publisher       = {{AAAI} Press},
  _url             = {https://ojs.aaai.org/index.php/ICAPS/article/view/6665},
  author           = {Enrico Scala and Alessandro Saetti and Ivan Serina and
                      Alfonso Emilio Gerevini},
  bibsource        = {dblp computer science bibliography, https://dblp.org},
  biburl           = {https://dblp.org/rec/conf/aips/ScalaSSG20.bib},
  booktitle        = {ICAPS},
  modificationdate = {2025-04-16T11:08:29},
  timestamp        = {Mon, 07 Mar 2022 16:58:36 +0100},
  title            = {Search-Guidance Mechanisms for Numeric Planning Through
                      Subgoaling Relaxation},
  year             = {2020},
}

@inproceedings{segoviaaguas.etal.2021,
  _note            = {\gp},
  _pages           = {569--577},
  author           = {Javier Segovia-Aguas and Sergio Jim{\'{e}}nez and Anders
                      Jonsson},
  booktitle        = {{ICAPS}},
  modificationdate = {2025-04-21T20:48:16},
  title            = {Generalized Planning as Heuristic Search},
  year             = {2021},
}

@inproceedings{shen.etal.2020,
  _note            = {\lp},
  _pages           = {574--584},
  author           = {Shen, William and Trevizan, Felipe and Thi{\'e}baux,
                      Sylvie},
  booktitle        = {ICAPS},
  modificationdate = {2025-03-26T14:40:27},
  title            = {{L}earning {D}omain-{I}ndependent {P}lanning {H}euristics
                      with {H}ypergraph {N}etworks},
  year             = {2020},
}

@inproceedings{sievers.etal.2019a,
  _editor          = {J. Benton and Nir Lipovetzky and Eva Onaindia and David E.
                      Smith and Siddharth Srivastava},
  _pages           = {446--454},
  _publisher       = {{AAAI} Press},
  _url             = {https://ojs.aaai.org/index.php/ICAPS/article/view/3509},
  author           = {Silvan Sievers and Gabriele R{\"{o}}ger and Martin Wehrle
                      and Michael Katz},
  bibsource        = {dblp computer science bibliography, https://dblp.org},
  biburl           = {https://dblp.org/rec/conf/aips/SieversRW019.bib},
  booktitle        = {ICAPS},
  modificationdate = {2025-03-29T14:03:04},
  timestamp        = {Tue, 20 Aug 2024 07:54:44 +0200},
  title            = {Theoretical Foundations for Structural Symmetries of
                      Lifted {PDDL} Tasks},
  year             = {2019},
}

@inproceedings{silver.etal.2024,
  _note            = {\gp},
  _pages           = {20256--20264},
  author           = {Tom Silver and Soham Dan and Kavitha Srinivas and Joshua
                      B. Tenenbaum and Leslie Kaelbling and Michael Katz},
  booktitle        = {AAAI},
  modificationdate = {2025-03-26T14:40:27},
  title            = {Generalized Planning in {PDDL} Domains with Pretrained Large
                      Language Models},
  year             = {2024},
}

@inproceedings{speck.etal.2019,
  _editor          = {Lipovetzky, Nir and Onaindia, Eva and Smith, David E.},
  _pages           = {464--572},
  _publisher       = {AAAI Press},
  author           = {Speck, David and Gei{\ss}er, Florian and Mattm{\"u}ller,
                      Robert and Torralba, {\'{A}lvaro}},
  booktitle        = {ICAPS},
  modificationdate = {2025-03-28T16:07:46},
  title            = {Symbolic Planning with Axioms},
  year             = {2019},
}

@inproceedings{srivastava.etal.2008,
  _note            = {\gp},
  _pages           = {991--997},
  author           = {Siddharth Srivastava and Neil Immerman and Shlomo
                      Zilberstein},
  booktitle        = {{AAAI}},
  modificationdate = {2025-03-26T14:40:27},
  title            = {Learning Generalized Plans Using Abstract Counting},
  year             = {2008},
}

@inproceedings{srivastava.etal.2011,
  _note            = {\gp},
  _pages           = {1010--1016},
  author           = {Siddharth Srivastava and Shlomo Zilberstein and Neil
                      Immerman and Hector Geffner},
  booktitle        = {{AAAI}},
  modificationdate = {2025-03-26T14:40:27},
  title            = {Qualitative Numeric Planning},
  year             = {2011},
}

@inproceedings{staahlberg.etal.2022,
  _note            = {\lp},
  _pages           = {629--637},
  author           = {Simon St{\aa}hlberg and Blai Bonet and Hector Geffner},
  booktitle        = {ICAPS},
  modificationdate = {2025-03-26T14:40:27},
  title            = {Learning General Optimal Policies with Graph Neural
                      Networks: Expressive Power, Transparency, and Limits},
  year             = {2022},
}

@inproceedings{toyer.etal.2018,
  _note            = {\lp},
  _pages           = {6294--6301},
  author           = {Sam Toyer and Felipe W. Trevizan and Sylvie Thi{\'{e}}baux
                      and Lexing Xie},
  booktitle        = {{AAAI}},
  creationdate     = {2024-09-04T17:49:00},
  modificationdate = {2025-03-26T14:40:27},
  title            = {Action Schema Networks: Generalised Policies With Deep
                      Learning},
  year             = {2018},
}

@inproceedings{yang.etal.2022,
  _doi             = {10.24963/IJCAI.2022/650},
  _editor          = {Luc De Raedt},
  _pages           = {4686--4692},
  _publisher       = {ijcai.org},
  _url             = {https://doi.org/10.24963/ijcai.2022/650},
  author           = {Ryan Yang and Tom Silver and Aidan Curtis and Tom{\'{a}}s
                      Lozano{-}P{\'{e}}rez and Leslie Pack Kaelbling},
  bibsource        = {dblp computer science bibliography, https://dblp.org},
  biburl           = {https://dblp.org/rec/conf/ijcai/YangSCLK22.bib},
  booktitle        = {IJCAI},
  modificationdate = {2025-04-21T09:59:50},
  timestamp        = {Tue, 15 Oct 2024 16:43:28 +0200},
  title            = {{PG3:} Policy-Guided Planning for Generalized Policy
                      Generation},
  year             = {2022},
}

@misc{favorito.etal.2025,
  author           = {Marco Favorito and Francesco Fuggitti and Christian
                      Muise},
  modificationdate = {2025-03-28T14:52:25},
  note             = {Accessed from \url{https://github.com/ai-Planning/pddl}},
  title            = {pddl},
  year             = {2025},
}

@misc{hipp.2020,
  author           = {Dwayne Richard Hipp},
  modificationdate = {2025-03-28T15:02:00},
  note             = {Accessed from \url{https://www.sqlite.org/index.html}},
  title            = {SQLite},
  year             = {2020},
}

@misc{ups.2025,
  author           = {UPS},
  howpublished     = {Accessed from
                      \url{https://about.ups.com/content/dam/upsstories/images/our-impact/reporting/2024-UPS-GRI-Report.pdf}},
  modificationdate = {2025-04-29T09:24:50},
  title            = {Global Reporting Initiative},
  year             = {2025},
}

@phdthesis{sussman.1973,
  author           = {Gerald Jay Sussman},
  modificationdate = {2025-03-29T09:47:40},
  school           = {MIT},
  title            = {A Computational Model of Skill Acquisition},
  year             = {1973},
}

@techreport{mcdermott.etal.1998,
  author           = {Drew McDermott and Malik Ghallab and Adele E. Howe and
                      Craig A. Knoblock and Ashwin Ram and Manuela M. Veloso and
                      Daniel S. Weld and David E. Wilkins},
  modificationdate = {2025-03-26T14:41:09},
  title            = {{PDDL}-the planning domain definition language},
  year             = {1998},
}

@inproceedings{howel.etal.2004,
  author           = {Richard Howey and Derek Long and Maria Fox},
  bibsource        = {dblp computer science bibliography, https://dblp.org},
  booktitle        = {ICTAI},
  doi              = {10.1109/ICTAI.2004.120},
  modificationdate = {2025-07-20T14:51:01},
  title            = {{VAL:} Automatic Plan Validation, Continuous Effects and Mixed Initiative Planning Using {PDDL}},
  url              = {https://doi.org/10.1109/ICTAI.2004.120},
  year             = {2004},
}

@article{korf.1987,
  author           = {Richard E. Korf},
  bibsource        = {dblp computer science bibliography, https://dblp.org},
  doi              = {10.1016/0004-3702(87)90051-8},
  journal          = {Artif. Intell.},
  modificationdate = {2025-07-22T09:25:22},
  number           = {1},
  pages            = {65--88},
  title            = {Planning as Search: {A} Quantitative Approach},
  url              = {https://doi.org/10.1016/0004-3702(87)90051-8},
  volume           = {33},
  year             = {1987},
}

@inproceedings{liu.etal.2025,
  author    = {Xiaotian Liu and Ali Pesaranghader and Hanze Li and Punyaphat Sukcharoenchaikul and Jaehong Kim and Tanmana Sadhu and Hyejeong Jeon and Scott Sanner},
  booktitle = {ACL},
  title     = {Open-World Planning via Lifted Regression with LLM-Inferred Affordances for Embodied Agents},
  year      = {2025},
}

@inproceedings{davidson.garagnani.2002,
  author    = {Marina Davidson and Max Garagnani},
  booktitle = {Proc. of the 21st Workshop of the UK Planning and Scheduling Special Interest Group (PlanSIG 2002)},
  title     = {Pre-processing planning domains containing Language Axioms},
  year      = {2002},
}

@article{lozano.etal.1984,
  author  = {Lozano-Perez, Tomas and Mason, Matthew T and Taylor, Russell H},
  journal = {Int. J. Robot. Res.},
  number  = {1},
  pages   = {3--24},
  title   = {Automatic synthesis of fine-motion strategies for robots},
  volume  = {3},
  year    = {1984},
}

@inproceedings{kaelbling.lozano.2011,
  author    = {Leslie Pack Kaelbling and
               Tom{\'{a}}s Lozano{-}P{\'{e}}rez},
  booktitle = {Robotics Research},
  title     = {Pre-image Backchaining in Belief Space for Mobile Manipulation},
  year      = {2011},
}

@inproceedings{xu.etal.2019,
  author       = {Danfei Xu and
                  Roberto Mart{\'{\i}}n{-}Mart{\'{\i}}n and
                  De{-}An Huang and
                  Yuke Zhu and
                  Silvio Savarese and
                  Li Fei{-}Fei},
  title        = {Regression Planning Networks},
  booktitle    = {NeurIPS},
  year         = {2019},
}
\normalsize

\clearpage
\onecolumn
\appendix

\section{Numeric Planning Definitions}\label{app:sec:numeric-background}
We can extend the definition of a planning problem to handle numeric variables, numeric conditions and numeric effects by making use of the fragment of PDDL 2.1~\cite{fox.long.2003} excluding durative actions.
In this paper, we consider the fragment of numeric planning where numeric conditions are restricted to involve one numeric variable per formula with comparisons in $\set{\geq, >, =}$, and numeric action effects to additions by a constant value.
This fragment is very expressive as it is undecidable by reduction from an abacus program~\cite[Theorem 12]{helmert.2002}.

\newcommand{\factsInS}{\ensuremath{X_p}}
\newcommand{\fluentsInS}{\ensuremath{X_n}}
\newcommand{\varsInState}{\ensuremath{X}}
\newcommand{\symbols}{\Sigma}
\newcommand{\symb}{\sigma}
\renewcommand{\arity}{n}
\renewcommand{\objects}{\mathcal{O}}
\renewcommand{\schemata}{\mathcal{A}}
\newcommand{\liftedProblem}{\gen{\predicates, \functions, \objects, \schemata, s_0, g}}

\begin{definition}[Abacus Program Numeric Planning Problem]\label{defn:num-problem}
  A (lifted) numeric planning problem is a tuple $\Pi = \liftedProblem$ where $\predicates$ is a set of lifted predicates, $\functions$ is a set of lifted (numeric) functions, $\objects$ is a set of objects, $\schemata$ is a set of action schemata, $s_0$ is the initial state, and $g$ the goal condition.
  Predicates are defined as in STRIPS planning and functions are similar where each function $f \in \functions$ has a set of argument terms $x_1, \ldots, x_{n_f}$ where $n_f \in \N_0$ depends on $f$.
  A propositional variable is a predicate whose argument terms are all instantiated with objects, and has a range in $\set{\top, \bot}$.
  A numeric variable is a function whose argument terms are all instantiated with objects, and has a range in $\R$.
  We let $N_p/N_n$ denote the set of propositional/numeric variables for $\Pi$ given by all possible instantiations of predicates/functions.
  A state is a value assignment to all $N_p$ and $N_n$.

  A literal is a predicate $p$ or its negation $\neg p$.
  A propositional condition is a positive (resp.\ negative) literal $x=\top$ (resp.\ $\bot$), and a numeric condition has the form $f \unrhd c$ where $f$ is a function, $c \in \R$ and $\unrhd \in \set{\geq, >, =}$.
  A state $s$ satisfies a set of conditions (i.e.\ a set of propositional and numeric conditions) if each condition in the set evaluates to true given the values of the state variables in $s$.
  The goal $g$ is a set of conditions.

  An action schema $a \in \schemata$ is a tuple $\gen{\var(a), \pre(a), \add(a), \del(a), \numEff(a)}$ where $\var(a)$ is a set of parameter terms, the preconditions $\pre(a)$ is a set of conditions, $\add(a)$ and $\del(a)$ are add and delete lists of predicates as in the classical case, $\numEff(a)$ is a set of numeric conditions of the form $f(x_1, \ldots, x_{n_f}) = f(x_1, \ldots, x_{n_f}) + c$ for $c \in \R$ with argument terms instantiated with terms or objects from $\var(a) \cup \objects$.
  An action $a$ is applicable in a state $s$ if $s$ satisfies $\pre(a)$.
  In this case, its successor $\succ(s, a)$ is the state where the effects $\numEff(a)$ are applied to the numeric variables in $s$, and propositional variables are modified in the same way as in the classical case.
  If $a$ is not applicable in $s$, we have $\succ(s, a) = \bot$.
  The definition of plan is the same as in the classical case.
\end{definition}

Next, we extend the definition of logical regression for classical STRIPS planning to handle numeric conditions and effects for the class of abacus program numeric problems.
\begin{definition}[Abacus Program Numeric Planning Regression]
  A set of conditions $g$ is regressable over an action $a$ if the classical conditions are regressable over $a$ as for STRIPS regression, with no restriction for numeric conditions.
  In this case, we define the regression of $g$ under $a$ by transforming numeric conditions $f \unrhd c$ to $f \unrhd c - v$ if there is an effect of $a$ of the form $f = f + v$, and if there exists any precondition $\xi = (f \unrhd c)$ in $a$, the associated numeric condition corresponding to $f$ in $g$ is transformed to $\xi$.
\end{definition}

\section{Proofs of Theorems}\label{app:sec:proofs}
This section proves theorems stated in \Cref{sec:theory}.

\subsection{Complexity Proofs}

\begin{proof}[Proof Sketch for \Cref{thm:tgi}]
  Membership follows from \PSPACE{}-completeness of planning for fixed domains~\cite{bylander.1994,erol.etal.1995}, and
  hardness by reduction from the Rush Hour problem which has singleton goals and has been shown to be \PSPACE{}-hard~\cite{hearn.demaine.2005}.
\end{proof}

\begin{proof}[Proof Sketch for \Cref{thm:sgi}]
  Membership again follows from \PSPACE{}-completeness of planning for fixed domains.
  Hardness follows from \Cref{thm:tgi} as \tgi{} is a special case of \sgi{}.
\end{proof}

\begin{proof}[Proof Sketch for \Cref{thm:ogi}]
  Membership again follows from \PSPACE{}-completeness of planning for fixed domains.
  Hardness follows from \Cref{thm:sgi} as \sgi{} is a special case of \ogi{}.
\end{proof}

\begin{proof}[Proof Sketch for \Cref{thm:ptgi}]
  This follows by noting that the greedy algorithm described in the definition of \tgi{} runs in polynomial time under the assumption that step (2a) can run in polynomial time.
\end{proof}

\begin{proof}[Proof Sketch for \Cref{thm:psgi}]
  For \NP{} membership, one guesses the correct ordering of goals after which running the greedy algorithm is in polynomial time.
  For \NP{}-hardness, we reduce from the Hamiltonian path problem which is \NP{}-complete~\cite[p. 60]{garey.johnson.1979}.
  The Hamiltonian path problem asks to find a path on a graph that visits every vertex exactly once.
  To encode this as a planning domain, one would require goals of the form $\textit{visited}(x)$ and actions which traverse a graph but make each vertex untraversable once it has been visited, such as by deleting initially true $\textit{clear}(x)$ atoms.
  Then the Hamiltonian path problem is equivalent to finding an correct ordering of goals representing (adjacent) vertices to visit.
\end{proof}

\begin{proof}[Proof Sketch for \Cref{thm:pogi}]
  For \NP{} membership, one guesses the correct ordering of goals after which running the greedy algorithm is in \NP{} time.
  For \NP{}-hardness, this follows from the \NP{}-hardness of \psgi{} which is a special case of \pogi{}.
\end{proof}

\subsection{Soundness and Completeness Proofs}

\begin{proof}[Proof Sketch for \Cref{thm:equivalence}]
  Reflexivity, symmetry and transitivity follows from the usage of bijective functions in the definition of $\equivalent$.
\end{proof}

\begin{proof}[Proof Sketch for \Cref{thm:equiv-plan}]
  The statement follows by definitions and bijectiveness of $f$.
\end{proof}

\newcommand{\abound}{\sum_{k=0}^C (\abs{\schemata}\cdot (kN'+M')^N)^k}
\newcommand{\gbound}{\abs{\predicates} \cdot M'^{M}}
\newcommand{\bound}{\ensuremath{\abound \cdot (\gbound)}}

\begin{proof}[Proof Sketch for \Cref{thm:sat}]
  We need to show that there exists a set $\trainProblems$ such that a \moose{} plan synthesised via \Cref{alg:learn} from $\trainProblems$ can solve any generalised planning problem $\gproblem = \gptuple$ exhibiting \btgi{}.
  To show completeness, it suffices to prove that there exists a finite number of \moose{} rules $\pi$, representing all possible optimal plans for singleton goals in step (2a) of the definition of TGI in \Cref{defn:tgi}, which when executed with \Cref{alg:plan} can solve any problem in $\gproblem$ with singleton goals.
  Then by the \tgi{} assumption, any arbitrary problem $\problem$ in $\problems$ can be solved because execution of $\pi$ would achieve the goals $\problem[g]$ in a monotonic fashion.
  Soundness follows by the fact that application of rules lifted from regression.
  Thus, the execution of \Cref{alg:plan} is sound as if a plan exists and is returned, then the execution of actions in sequence is valid and achieves the goal, and otherwise if no plan exists, then the algorithm would terminate and not return a plan.

  Now, note that the set of possible singleton goals in $\gproblem$ is finite modulo the equivalence relation $\equivalent$ restricted to sets of atoms, as there are a finite number of predicates and instantiations of nonequivalent objects.
  Furthermore by the \btgi{} assumption any optimal plan for any possible singleton goal $g$ has plan length less than $C$.
  Thus the set of all possibles sequences of actions that can be regressed from $g$ is bounded by $\abound$ where
  \begin{itemize}
    \item $N = \max_{a \in \schemata}(\abs{var(a)})$ is the maximum arity of schemata,
    \item $M$ is the maximum arity of predicates,
    \item $N' = N + \abs{\constants}$, and
    \item $M' = M + \abs{\constants}$.
  \end{itemize}
  Note that by \Cref{thm:equiv-plan}, it suffices to count the equivalence class of plans under $\equivalent$.
  This is because modulo $\equivalent$ there are at most $({kN'+M'})^N$ possible instantiations of an action where there are at most $kN'+M'$ possible objects across all actions in a length-$k$ plan and the singleton goal under $\equivalent$.
  By a similar argument, there are $\gbound$ possible singleton goal instantiations, and hence it takes a finite number of up to $n=\bound$ different problems to synthesise a general policy for $\GP$.
\end{proof}

\begin{proof}[Proof Sketch for \Cref{thm:ogi}]
  \NP{} membership follows by definition of \ogi{}, more specifically the nondeterministic algorithm defined within.
  \NP{}-hardness follows by the \NP{}-hardness of \sgi{} which can be viewed as a general case of \ogi{}.
\end{proof}

\begin{proof}[Proof Sketch for \Cref{thm:opt}]
  Completeness follows from the completeness of rules as discussed in \Cref{thm:sat}.
  To show soundness for optimal planning, we note that given enough training problems, bounded by $n = \bound$ from the previous theorem, learned rules do not throw away any optimal actions at every state.
  Then the proof follows from the definition of \ogi{}.
\end{proof}

\begin{figure}[h!]
  \centering
  \newcommand{\angl}{10}
  \newcommand{\xshift}{3}
  \newcommand{\xshiftb}{\xshift*1.32}
  \newcommand{\Size}{0.6cm}

\begin{tikzpicture}
  \tikzset{
    node/.style={
      draw,
      rectangle,
      rounded corners,
      minimum size=\Size,
  }}
  \node[node] (A) at (0,0) {};

  \node[node] (B) at ($(A) + (+\angl:\xshift)$) {};
  \node[node] (C) at ($(B) + (+\angl:\xshift)$) {$g_1$};
  \node[node] (D) at ($(C) + (-\angl:\xshift)$) {};

  \node[node] (F) at ($(A) + (-\angl:\xshift)$) {};
  \node[node] (G) at ($(F) + (-\angl:\xshift)$) {$g_2$};
  \node[node] (H) at ($(G) + (+\angl:\xshift)$) {};

  \node[node] (I) at ($(A) + (0:\xshiftb)$) {};
  \node[node] (J) at ($(I) + (0:\xshiftb)$) {};

  \node[node] (e) at ($(D) + (-\angl:\xshift)$) {\color{white}$g_1, g_2$};
  \node[node, minimum size = 1.4*\Size] (E) at (e) {$\;g_1, g_2\;$};

  \draw[->] ($(A) + (180:0.3*\xshift)$) -- (A);

  \draw[->] (A) -- (B);
  \draw[->] (B) -- (C);
  \draw[->] (C) -- (D);
  \draw[->] (D) -- (E);

  \draw[->] (A) -- (F);
  \draw[->] (F) -- (G);
  \draw[->] (G) -- (H);
  \draw[->] (H) -- (E);

  \draw[->] (A) -- (I);
  \draw[->] (I) -- (J);
  \draw[->] (J) -- (E);
\end{tikzpicture}
  \caption{A planning problem illustrating the necessity of the \ogi{} assumption for learning provably optimal policies in \Cref{thm:opt}. Nodes represent states and arrows represent transitions between states.}
  \label{fig:counterexample}
\end{figure}
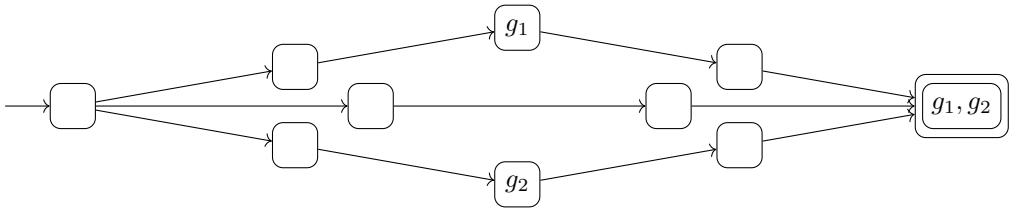

\begin{example}[\Cref{thm:opt} counterexample without the OGI assumption]\label{eg:ceg}
  A counterexample to the previous theorem for when the \ogi{} assumption is dropped occurs if we can find a case where achieving a singleton goal for a \tgi{} problem suboptimally is necessary to achieve optimality for the whole problem.
  In the planning problem illustrated by the state space in \Cref{fig:counterexample} with goal $\set{g_1, g_2}$, an optimal plan to either $g_1$ or $g_2$ has plan length 2 and the greedy algorithm in \Cref{defn:tgi} of \tgi{} returns a plan of length 4.
  However, the optimal plan has length 3.
\end{example}

\clearpage
\section{True Goal Independence Experiments}\label{app:sec:tgi}
Results for the Agricola domain are omitted as no problem could be solved in the given resource limits.
\begin{table}[h!]
  \begin{tabularx}{\textwidth}{l Y Y Y Y}
\toprule
Domain & OOR & Inval & Val & Val (\%) \\
\midrule
barman & 1 & 23 & 16 & 41.0 \\
blocks & 0 & 35 & 0 & 0.0 \\
childsnack & 5 & 0 & 15 & 100.0 \\
data & 15 & 0 & 5 & 100.0 \\
depot & 1 & 14 & 7 & 33.3 \\
driverlog & 0 & 18 & 2 & 10.0 \\
elevators & 10 & 0 & 40 & 100.0 \\
floortile & 0 & 40 & 0 & 0.0 \\
freecell & 31 & 0 & 49 & 100.0 \\
ged & 1 & 19 & 0 & 0.0 \\
grid & 0 & 3 & 2 & 40.0 \\
gripper & 0 & 0 & 20 & 100.0 \\
hiking & 17 & 0 & 3 & 100.0 \\
logistics & 5 & 0 & 58 & 100.0 \\
miconic & 0 & 0 & 150 & 100.0 \\
movie & 0 & 30 & 0 & 0.0 \\
mprime & 11 & 0 & 24 & 100.0 \\
mystery & 8 & 4 & 18 & 81.8 \\
nomystery & 0 & 20 & 0 & 0.0 \\
parking & 0 & 40 & 0 & 0.0 \\
pegsol & 0 & 50 & 0 & 0.0 \\
pipesworld & 37 & 51 & 12 & 19.1 \\
rovers & 0 & 0 & 40 & 100.0 \\
satellite & 6 & 15 & 15 & 50.0 \\
scanalyzer & 0 & 46 & 4 & 8.0 \\
snake & 0 & 20 & 0 & 0.0 \\
sokoban & 0 & 48 & 2 & 4.0 \\
spider & 11 & 9 & 0 & 0.0 \\
storage & 0 & 27 & 3 & 10.0 \\
termes & 17 & 3 & 0 & 0.0 \\
tetris & 0 & 20 & 0 & 0.0 \\
thoughtful & 15 & 1 & 4 & 80.0 \\
tidybot & 1 & 2 & 17 & 89.5 \\
tpp & 8 & 0 & 22 & 100.0 \\
transport & 44 & 0 & 26 & 100.0 \\
visitall & 18 & 0 & 22 & 100.0 \\
woodworking & 0 & 48 & 2 & 4.0 \\
zenotravel & 0 & 8 & 12 & 60.0 \\
\bottomrule
\end{tabularx}

  \caption{
    Results for the procedure described in \ref{q:tgi} in \Cref{sec:experiments}.
    OOR denotes that planning did not complete in the given resources.
    Inval denotes that a returned plan is invalid or that the procedure encountered a deadend.
    Val denotes that a returned plan is valid.
    Val (\%) is computed by $100\frac{\text{Val}}{\text{Val} + \text{Inval}}$.
  }
\end{table}

\clearpage
\section{Additional Generalised Planning Benchmark Details}\label{app:sec:benchmarks}
\Cref{tab:objects} displays the ranges of number of objects present in training and testing problems for each domain.
The Ferry, Miconic, Rovers, Satellite, and Transport domains and problems are taken from the 2023 IPC Learning Track \cite{taitler.etal.2024} with modifications that remove the path-finding components of Rovers and Transport, and that increase the number of objects in Miconic problems.
Barman, Gripper and Logistics have newly introduced training and testing problem splits for generalised planning.
The NFerry, NMiconic, and NTransport domains are taken from \cite{chen.thiebaux.2024} with path-finding removed from NTransport.
NMinecraft originates from the Minecraft Pogo Stick domain~\cite{benyamin.etal.2024} while we introduced training and testing problem splits for generalised planning.

\newcommand{\dwidth}{0.49\textwidth}

\begin{table}[h!]
  \scriptsize
  \tabcolsep 3pt
  \renewcommand{\arraystretch}{0.9}

  \newcommand{\splitter}{
    \cmidrule{2-2}
    \cmidrule(l){3-4}
    \cmidrule(l){5-6}
  }

  \begin{subtable}{\dwidth}
    \centering
    \begin{tabularx}{\textwidth}{X X Y Y Y Y}
  \toprule
  &
  &
  \multicolumn{2}{c}{Train} &
  \multicolumn{2}{c}{Test} \\
  \cmidrule(l){3-4}
  \cmidrule(l){5-6}
  {Domain} & {Obj. Types} & Min & Max & Min & Max \\
  \midrule
  Barman
  & $\Sigma$ & 16 & 27 & 21 & 853 \\
  \splitter
  & cocktail & 3 & 7 & 4 & 393 \\
  & dispenser & 3 & 3 & 3 & 30 \\
  & hand & 2 & 2 & 2 & 2 \\
  & ingredient & 3 & 3 & 3 & 30 \\
  & level & 3 & 3 & 3 & 3 \\
  & shaker & 1 & 1 & 1 & 1 \\
  & shot & 1 & 8 & 5 & 394 \\
  \midrule
  Ferry
  & $\Sigma$ & 3 & 8 & 7 & 1461 \\
  \splitter
  & car & 1 & 2 & 2 & 974 \\
  & location & 2 & 6 & 5 & 487 \\
  \midrule
  Gripper
  & $\Sigma$ & 3 & 5 & 11 & 48500 \\
  \splitter
  & balls & 3 & 5 & 11 & 48500 \\
  \midrule
  Logistics
  & $\Sigma$ & 29 & 29 & 10 & 1260 \\
  \splitter
  & airplane & 3 & 3 & 1 & 64 \\
  & city & 3 & 3 & 1 & 64 \\
  & location & 15 & 15 & 2 & 960 \\
  & package & 5 & 5 & 5 & 108 \\
  & truck & 3 & 3 & 1 & 64 \\
  \midrule
  Miconic
  & $\Sigma$ & 3 & 11 & 5 & 1950 \\
  \splitter
  & floor & 2 & 7 & 4 & 980 \\
  & passenger & 1 & 4 & 1 & 970 \\
  \midrule
  Rovers
  & $\Sigma$ & 10 & 36 & 12 & 596 \\
  \splitter
  & camera & 1 & 4 & 1 & 99 \\
  & lander & 1 & 1 & 1 & 1 \\
  & mode & 3 & 3 & 3 & 3 \\
  & objective & 1 & 10 & 1 & 236 \\
  & rover & 1 & 4 & 1 & 30 \\
  & store & 1 & 4 & 1 & 30 \\
  & waypoint & 2 & 10 & 4 & 197 \\
  \midrule
  Satellite
  & $\Sigma$ & 5 & 43 & 11 & 402 \\
  \splitter
  & direction & 2 & 10 & 4 & 98 \\
  & instrument & 1 & 20 & 3 & 195 \\
  & mode & 1 & 3 & 1 & 10 \\
  & satellite & 1 & 10 & 3 & 99 \\
  \midrule
  Transport
  & $\Sigma$ & 6 & 17 & 12 & 354 \\
  \splitter
  & location & 2 & 7 & 5 & 99 \\
  & package & 1 & 4 & 1 & 194 \\
  & size & 2 & 3 & 3 & 11 \\
  & vehicle & 1 & 3 & 3 & 50 \\
  \bottomrule
\end{tabularx}

    \caption{Classical planning.}
    \label{tab:classical}
  \end{subtable}
  \hfill
  \begin{subtable}{\dwidth}
    \centering
    \begin{tabularx}{\textwidth}{X X Y Y Y Y}
  \toprule
  &
  &
  \multicolumn{2}{c}{Train} &
  \multicolumn{2}{c}{Test} \\
  \cmidrule(l){3-4}
  \cmidrule(l){5-6}
  {Domain} & {Objects} & Min & Max & Min & Max \\
  \midrule
  NFerry
  & $\Sigma$ & 7 & 9 & 11 & 1465 \\
  \splitter
  & car & 1 & 2 & 2 & 974 \\
  & location & 2 & 3 & 5 & 487 \\
  \midrule
  NMiconic
  & $\Sigma$ & 7 & 15 & 9 & 685 \\
  \splitter
  & floor & 2 & 7 & 4 & 196 \\
  & passenger & 1 & 4 & 1 & 485 \\
  \midrule
  NMinecraft
  & $\Sigma$ & 5 & 5 & 15 & 2100 \\
  \splitter
  & cell & 4 & 4 & 11 & 1800 \\
  & pogo sticks & 1 & 1 & 4 & 300 \\
  \midrule
  NTransport
  & $\Sigma$ & 5 & 14 & 13 & 605 \\
  \splitter
  & capacity & 1 & 4 & 4 & 262 \\
  & location & 2 & 4 & 5 & 99 \\
  & package & 1 & 4 & 1 & 194 \\
  & vehicle & 1 & 2 & 3 & 50 \\
  \bottomrule
\end{tabularx}

    \caption{Numeric planning.}
    \label{tab:numeric}
  \end{subtable}
  \caption{Training and testing object distributions.}
  \label{tab:objects}
\end{table}

\clearpage
\section{Additional Generalised Planning Results}\label{app:sec:experiments}
\subsection{Coverage Tables}\label{app:ssec:coverage}
\begin{table*}[h!]
  \centering
  \renewcommand{\arraystretch}{1.2}
  \setlength{\tabcolsep}{1mm}
  \newcommand{\tabWidth}{0.33\textwidth}
  \small
  \begin{subtable}{\tabWidth}

\begin{tabularx}{\columnwidth}{l *{3}{Y}}
  Domain
  & \header{\enhspMQ{}}
  & \header{\enhspHMRPHJ{}}
  & \header{\moose{}}
  \\
  \cmidrule(l){1-1}
  \cmidrule(l){2-4}
  NFerry & 60 & 61 & \first{90.0} \\
  NMiconic & 63 & 71 & \first{90.0} \\
  NMinecraft & 30 & 66 & \first{90.0} \\
  NTransport & 44 & 64 & \first{90.0} \\
  \cmidrule(l){1-1}
  \cmidrule(l){2-4}
  $\sum$ (360)
 & 197 & 262 & \first{360.0} \\
\end{tabularx}
    \caption{Satisficing numeric planning.}
  \end{subtable}
  \begin{subtable}{\tabWidth}

\begin{tabularx}{\columnwidth}{l *{5}{Y}}
  Domain
  & \header{\slearnerA{}}
  & \header{\slearnerB{}}
  & \header{\slearnerC{}}
  & \header{\lama{}}
  & \header{\moose{}}
  \\
  \cmidrule(l){1-1}
  \cmidrule(l){2-6}
  Barman & 0.0 & 0.0 & 0.0 & 49 & \first{90.0} \\
  Ferry & 15.0 & 67.0 & 60.0 & 69 & \first{90.0} \\
  Gripper & 59.6 & 50.8 & 33.0 & 65 & \first{90.0} \\
  Logistics & 0.0 & 0.0 & 0.0 & 77 & \first{89.6} \\
  Miconic & 68.8 & 72.6 & 67.8 & 77 & \first{90.0} \\
  Rovers & 0.0 & 0.0 & 0.0 & 66 & \first{90.0} \\
  Satellite & 0.0 & 29.2 & 34.6 & 89 & \first{90.0} \\
  Transport & 0.0 & 63.0 & 46.8 & 66 & \first{90.0} \\
  \cmidrule(l){1-1}
  \cmidrule(l){2-6}
  $\sum$ (720)
 & 143.4 & 282.6 & 242.2 & 558 & \first{719.6} \\
\end{tabularx}
    \caption{Satisficing classical planning.}
  \end{subtable}
  \begin{subtable}{\tabWidth}

\begin{tabularx}{\columnwidth}{l *{5}{Y}}
  Domain
  & \header{\blind{}}
  & \header{\lmcut{}}
  & \header{\scorpion{}}
  & \header{\symk{}}
  & \header{\moose{}}
  \\
  \cmidrule(l){1-1}
  \cmidrule(l){2-6}
  Barman & 0 & 0 & 0 & 12 & \first{24.6} \\
  Ferry & 10 & 18 & 17 & 18 & \first{30.0} \\
  Gripper & 9 & 8 & 7 & \first{30} & 27.0 \\
  Logistics & 8 & 15 & \first{22} & 10 & 15.0 \\
  Miconic & \first{30} & \first{30} & \first{30} & \first{30} & \first{30.0} \\
  Rovers & 15 & 17 & 18 & \first{20} & \first{20.0} \\
  Satellite & 12 & 22 & \first{26} & 21 & 21.4 \\
  Transport & 9 & 9 & \first{20} & 13 & 15.0 \\
  \cmidrule(l){1-1}
  \cmidrule(l){2-6}
  $\sum$ (240)
 & 93 & 119 & 140 & 154 & \first{183.0} \\
\end{tabularx}
    \caption{Optimal classical planning.}
  \end{subtable}
  \caption{
    Planning coverage ($\uparrow$).
    The best score for each domain is highlighted in colour and bold.
    Domains have 90 (resp.\ 30) problems each for satisficing (resp.\ optimal) planning.
  }
  \label{tab:coverage}
\end{table*}

\newcommand{\insertLinePlot}[2]{
  \begin{figure}[h!]
    \centering
    \newcommand{\ccc}{0.37}
    \includegraphics[width=\ccc\textwidth]{./figures/line-plots/#1-time.pdf}
    \qquad
    \includegraphics[width=\ccc\textwidth]{./figures/line-plots/#1-plan_length.pdf}
    \caption{
      Average time (left) and plan length (right) of planners across solved problems for \textbf{#2}.
      Planning problem difficulty increases across the $x$-axis.
      Lower $y$-axis values are better ($\downarrow$).
      Note the $y$-axis log scale for runtime.
      Only points for which all seeds solve the problem are displayed.
    }
  \end{figure}
}

\subsection{Satisficing Numeric Planning}
\insertLinePlot{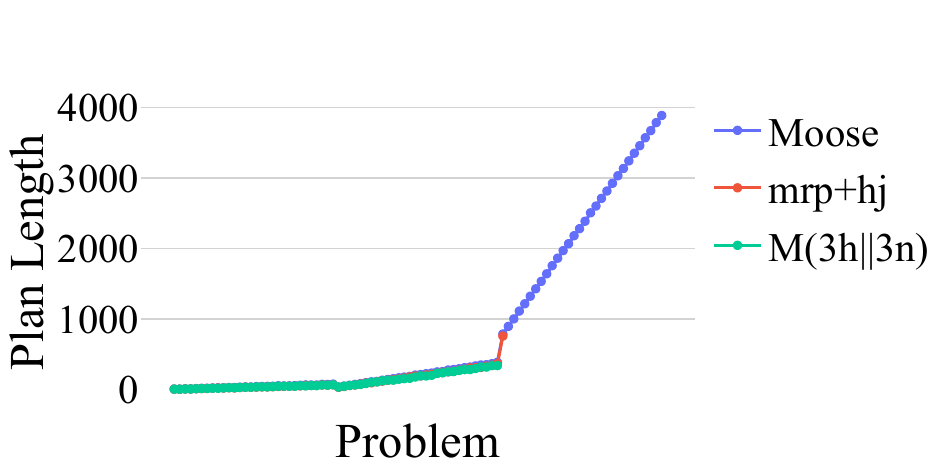}{Numeric Ferry}
\insertLinePlot{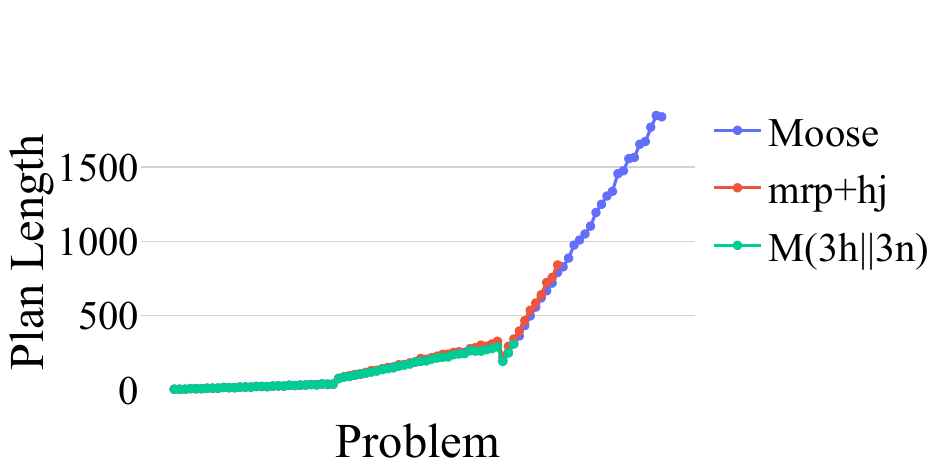}{Numeric Miconic}
\clearpage
\insertLinePlot{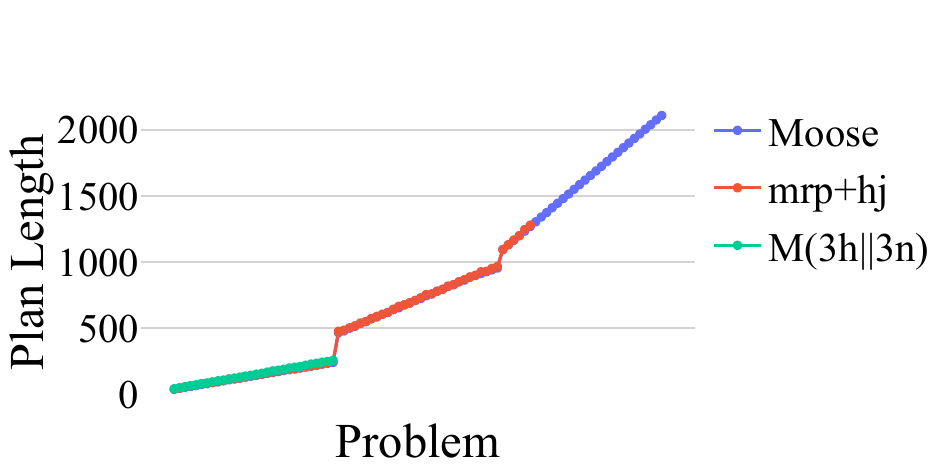}{Numeric Minecraft}
\insertLinePlot{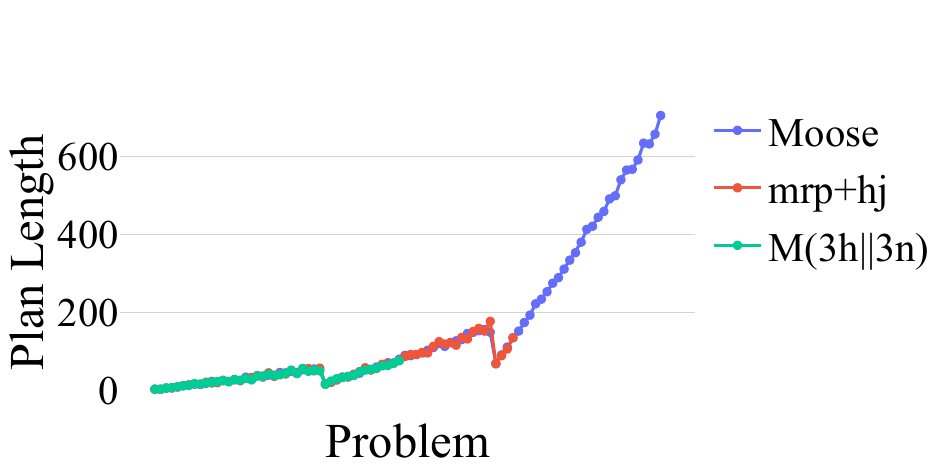}{Numeric Transport}

\subsection{Satisficing Classical Planning}
\insertLinePlot{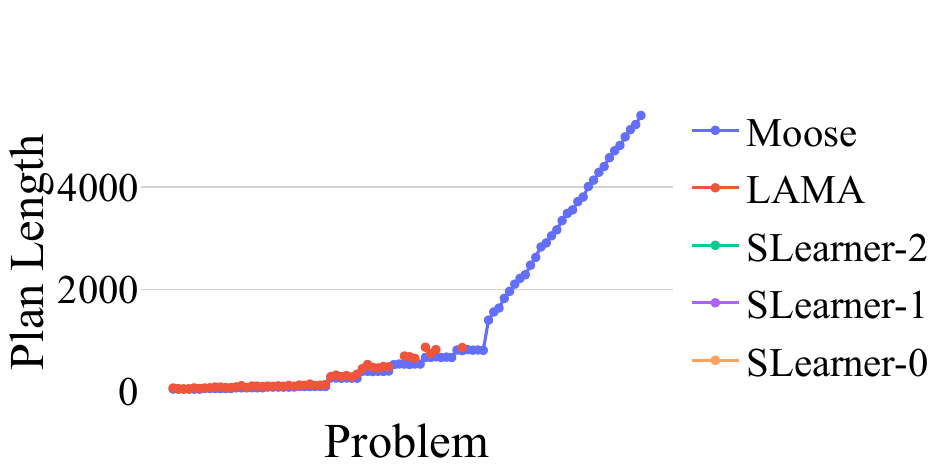}{Barman}
\insertLinePlot{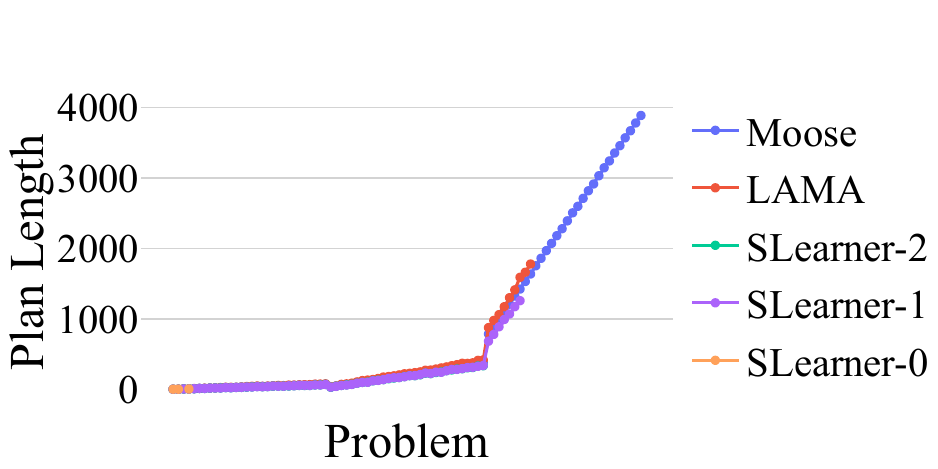}{Ferry}
\insertLinePlot{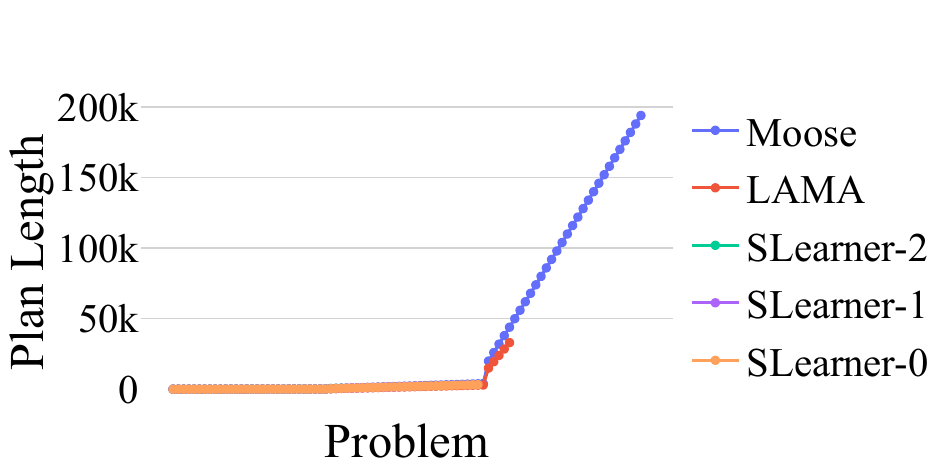}{Gripper}
\insertLinePlot{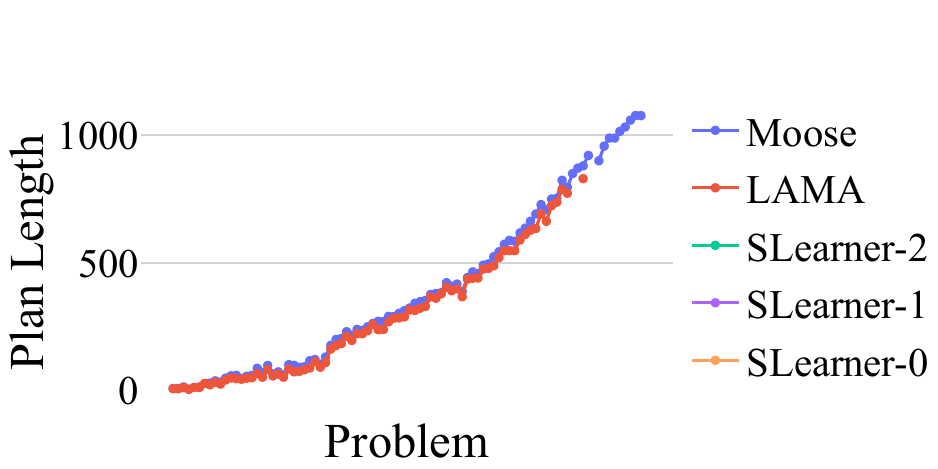}{Logistics}
\insertLinePlot{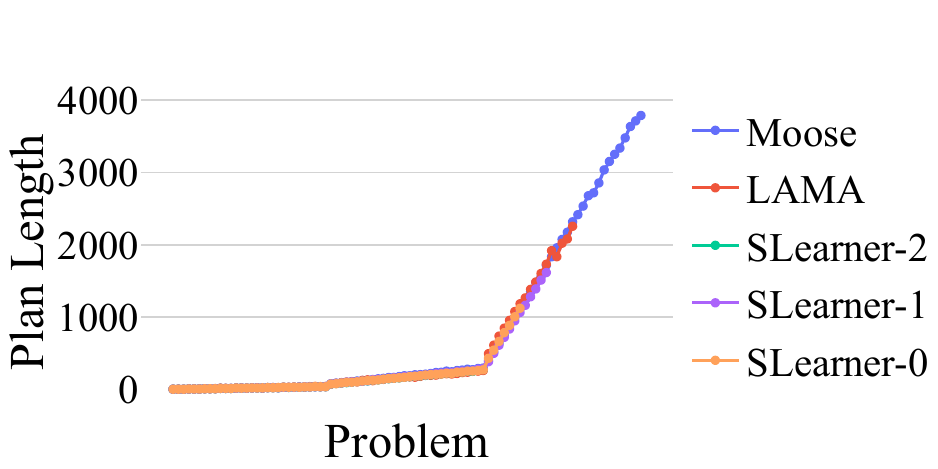}{Miconic}
\insertLinePlot{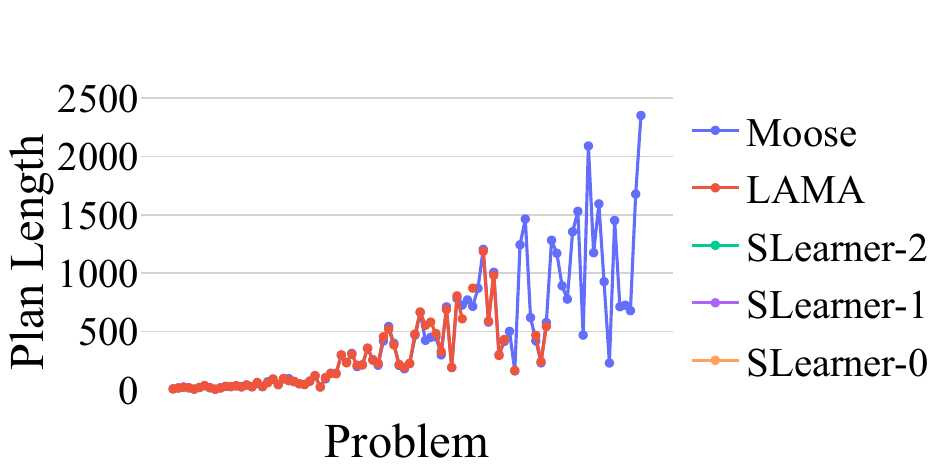}{Rovers}
\clearpage
\insertLinePlot{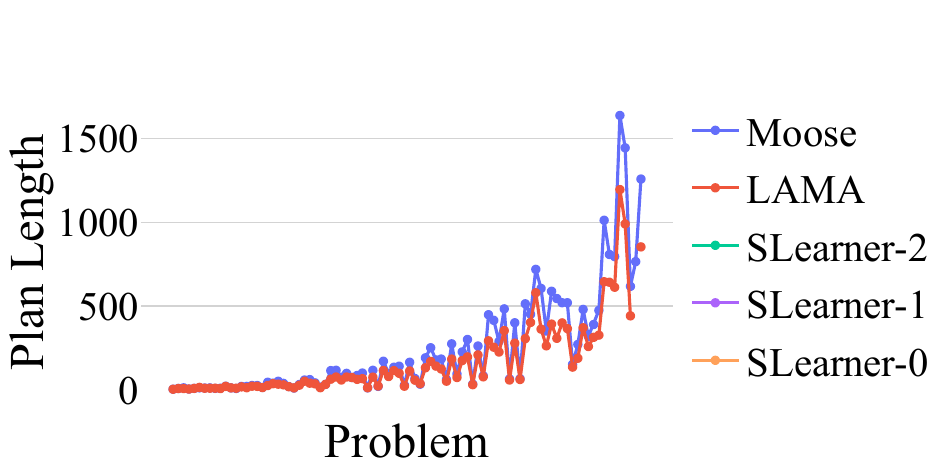}{Satellite}
\insertLinePlot{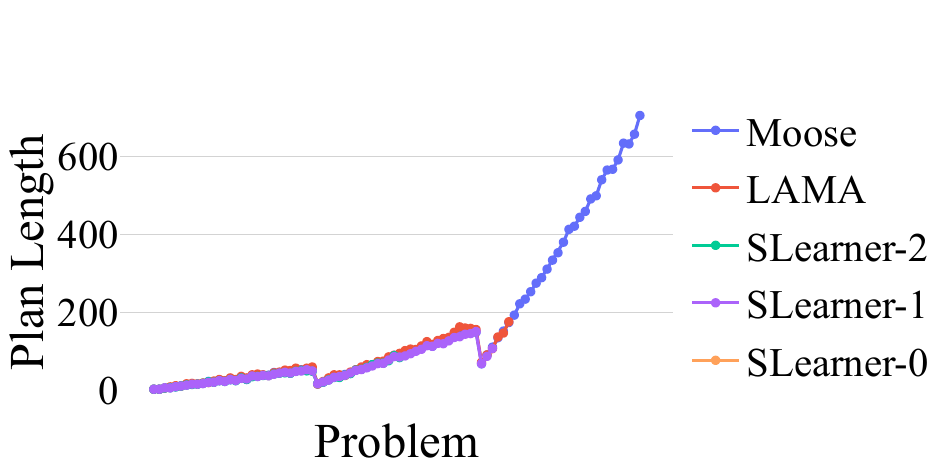}{Transport}

\subsection{Optimal Classical Planning}
\insertLinePlot{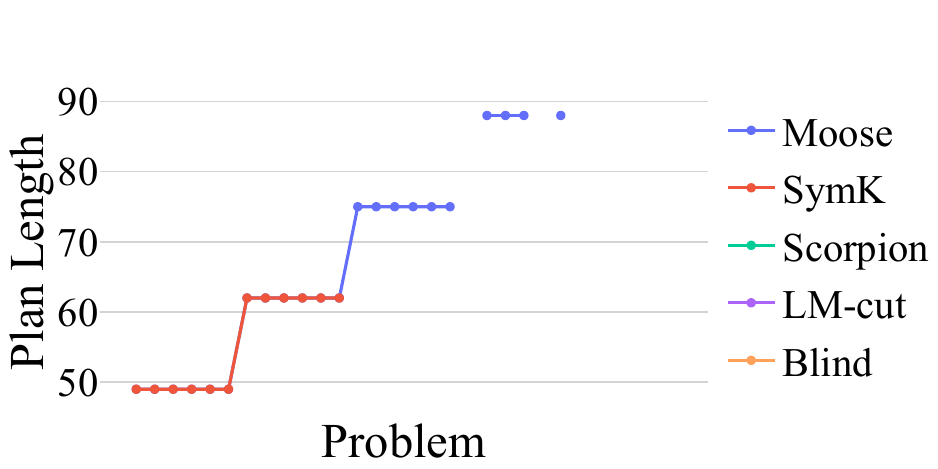}{Barman}
\insertLinePlot{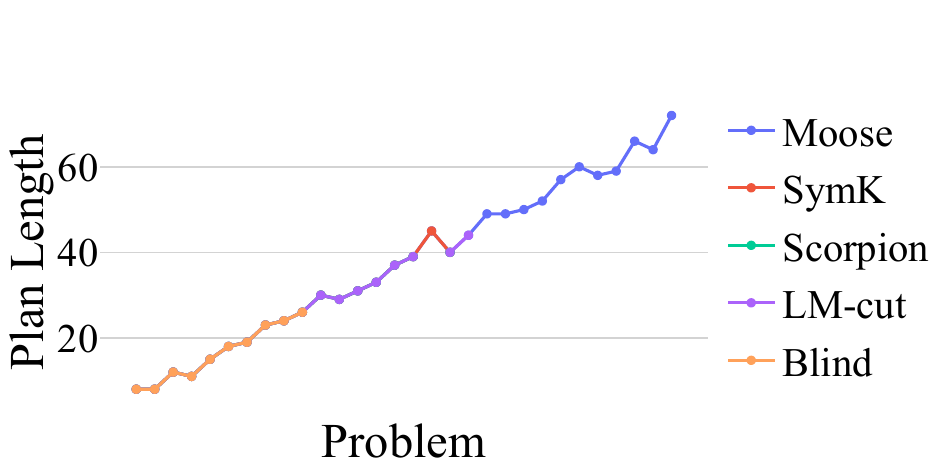}{Ferry}
\clearpage
\insertLinePlot{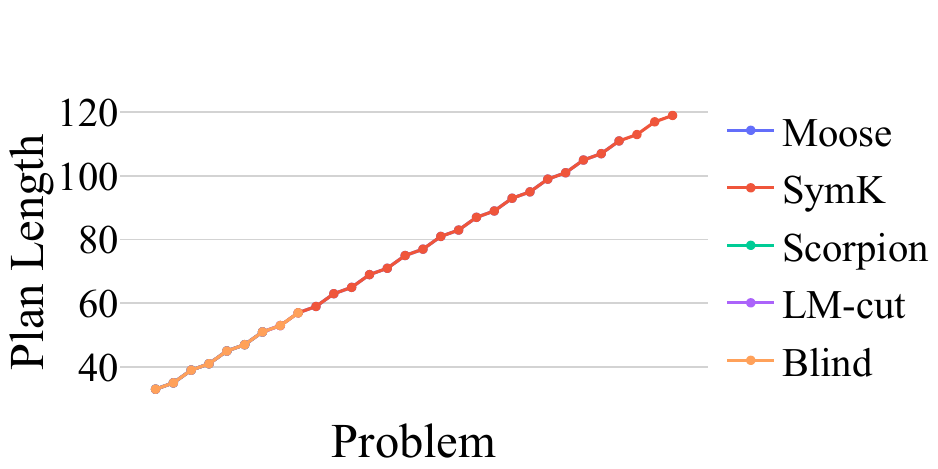}{Gripper}
\insertLinePlot{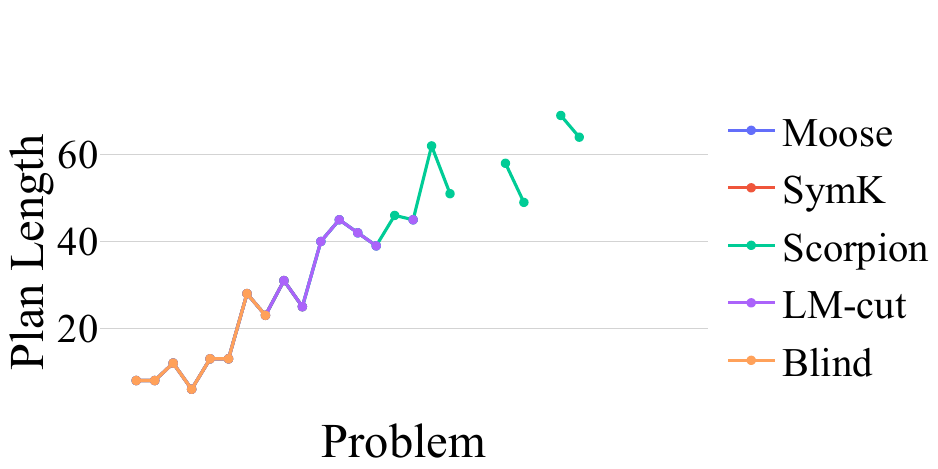}{Logistics}
\insertLinePlot{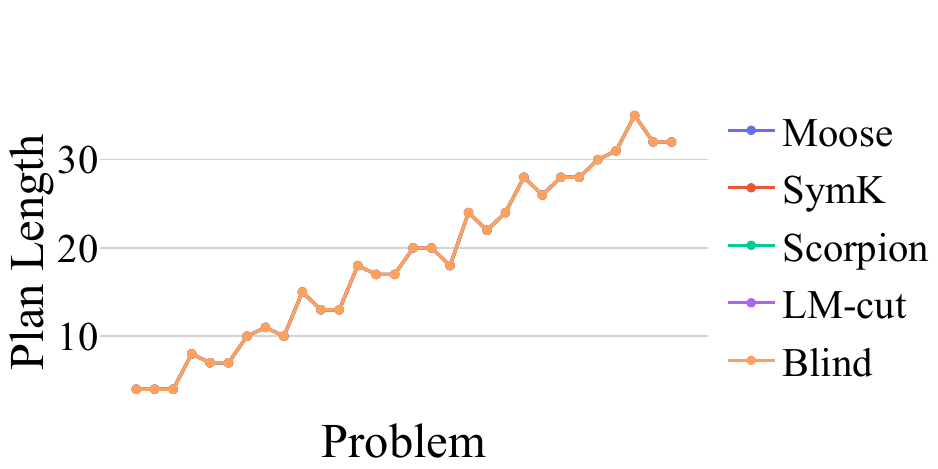}{Miconic}
\insertLinePlot{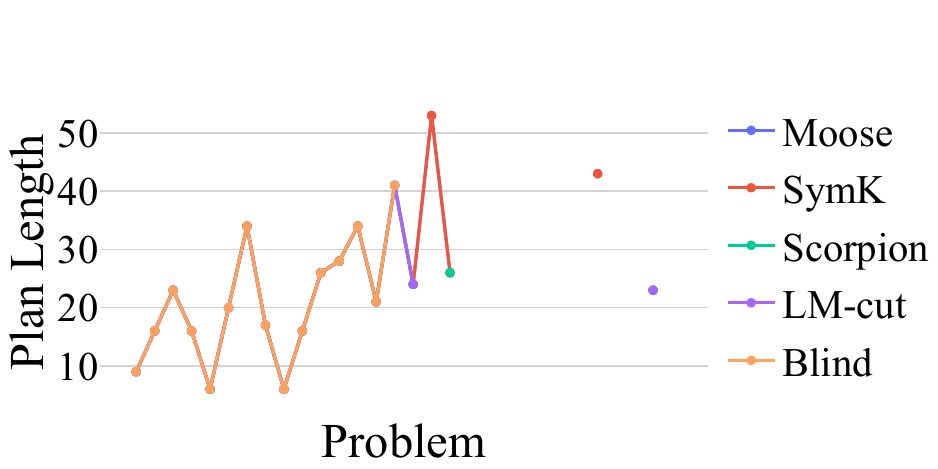}{Rovers}
\clearpage
\insertLinePlot{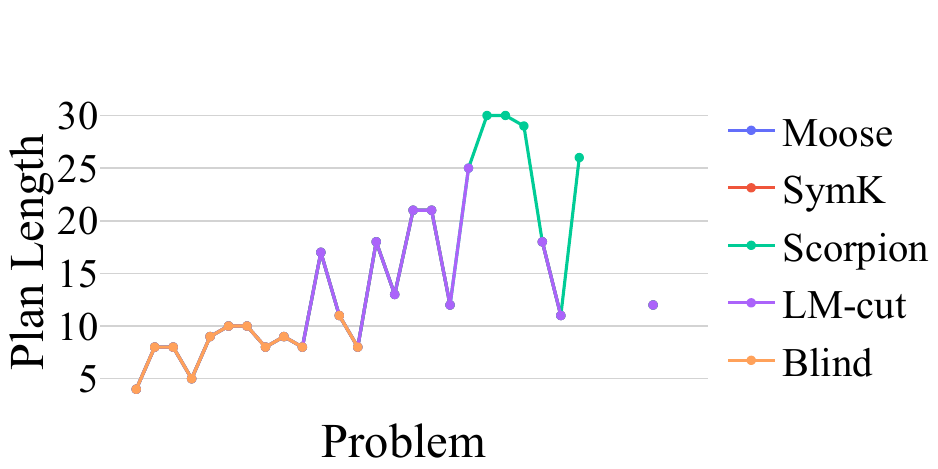}{Satellite}
\insertLinePlot{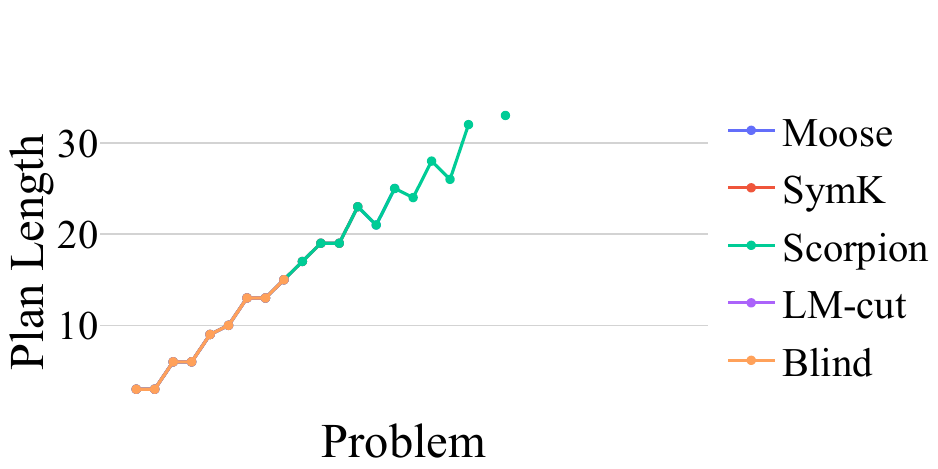}{Transport}

\subsection{\moose{} vs. \lama{} on Solution Quality}\label{app:ssec:versus-lama}
\begin{figure}[h!]
  \centering
  \newcommand{\ccc}{0.225}
  \includegraphics[width=\ccc\textwidth]{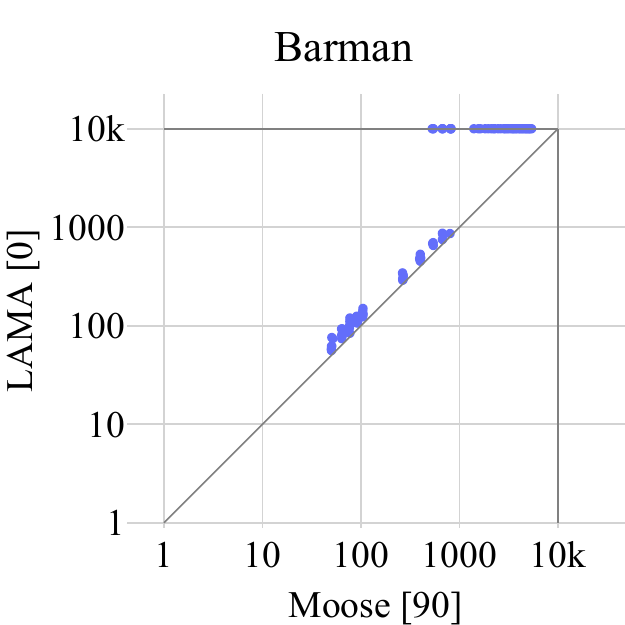}
  \includegraphics[width=\ccc\textwidth]{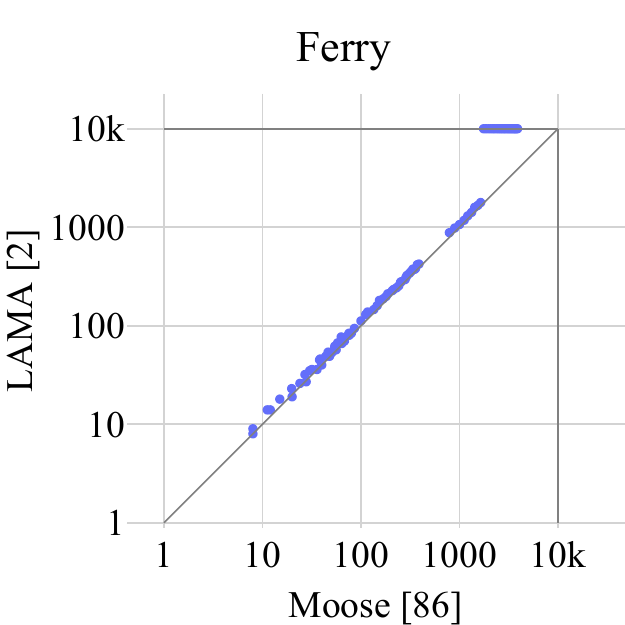}
  \includegraphics[width=\ccc\textwidth]{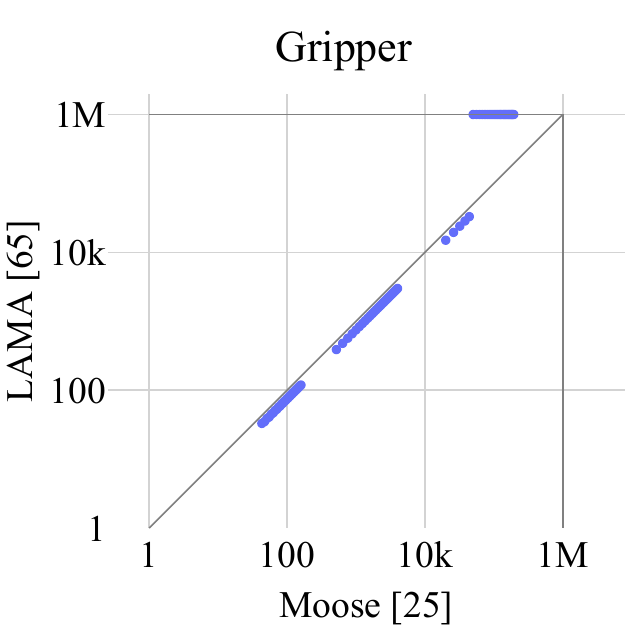}
  \includegraphics[width=\ccc\textwidth]{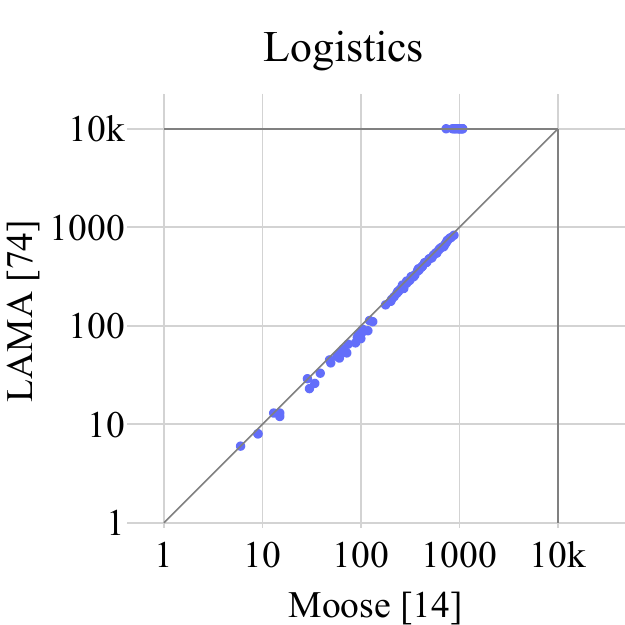}
  \includegraphics[width=\ccc\textwidth]{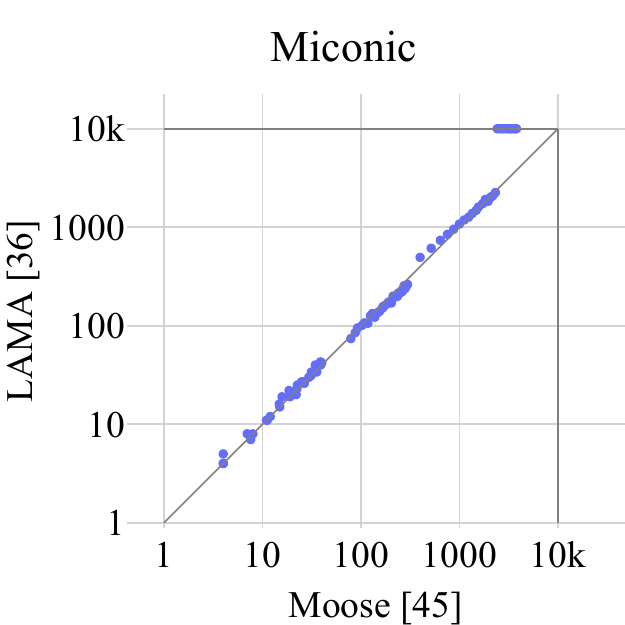}
  \includegraphics[width=\ccc\textwidth]{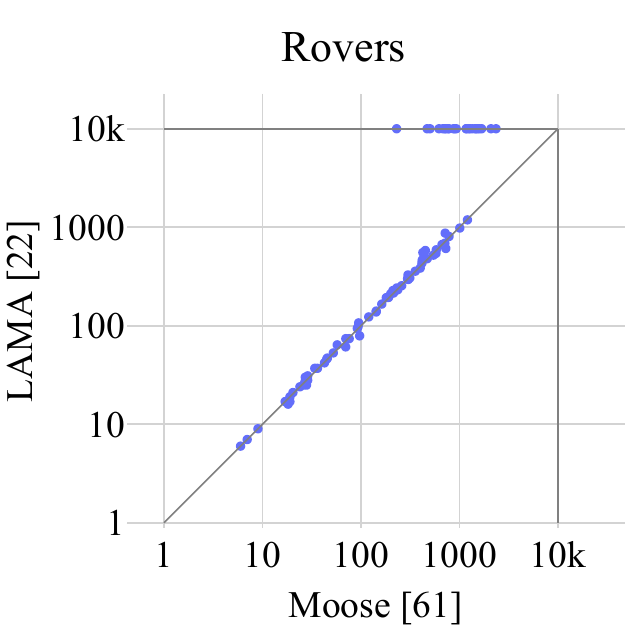}
  \includegraphics[width=\ccc\textwidth]{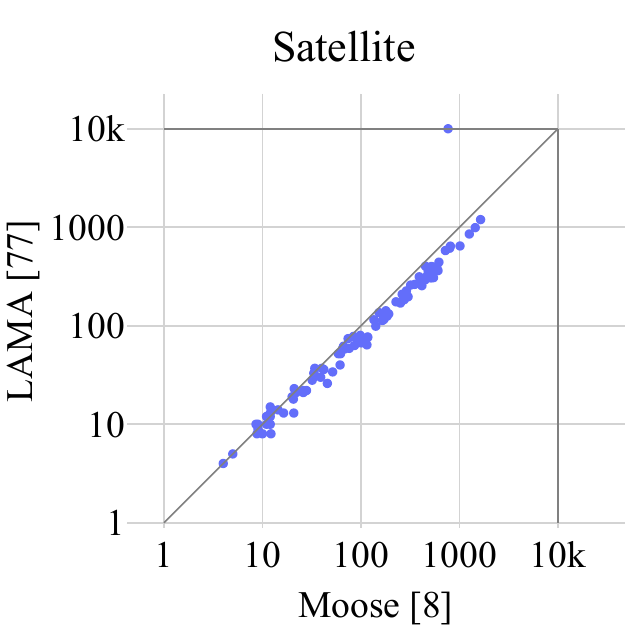}
  \includegraphics[width=\ccc\textwidth]{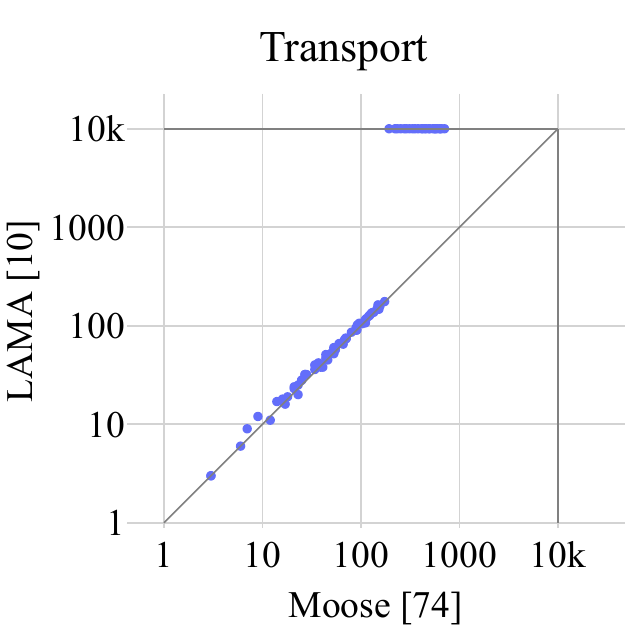}
  \caption{
    Plot comparisons of expanded nodes of \lama{} ($y$-axis) and \moose{} ($x$-axis) for different classical planning domains.
    A point $(x, y)$ represents the metric of the models indicated on the $x$ and $y$ axis on the domain.
    The number in the brackets next to each model indicates how many planning problems the model returned a higher quality plan than the model on the other axis.
    Points on the top left (resp.\ bottom right) triangle favour \moose{} (resp.\ \lama{}).
  }
  \label{fig:versus-lama}
\end{figure}

\subsection{\moose{} vs. \enhsp{} on Solution Quality}\label{app:ssec:versus-enhsp}
\begin{figure}[h!]
  \centering
  \newcommand{\ccc}{0.225}
  \includegraphics[width=\ccc\textwidth]{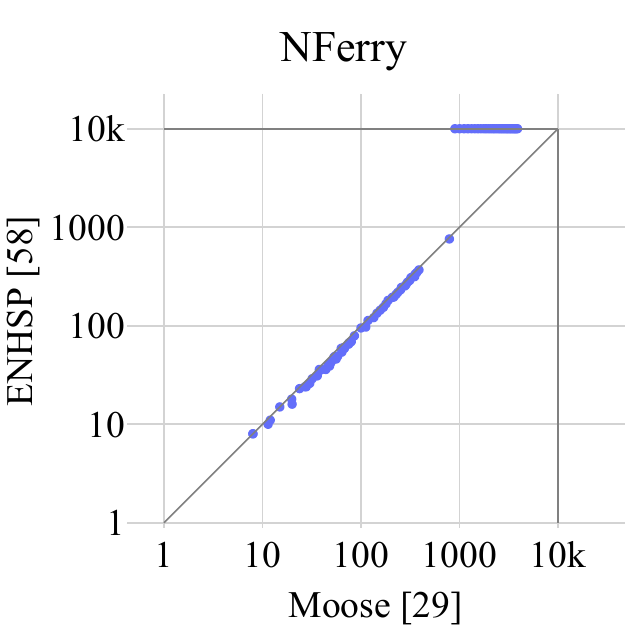}
  \includegraphics[width=\ccc\textwidth]{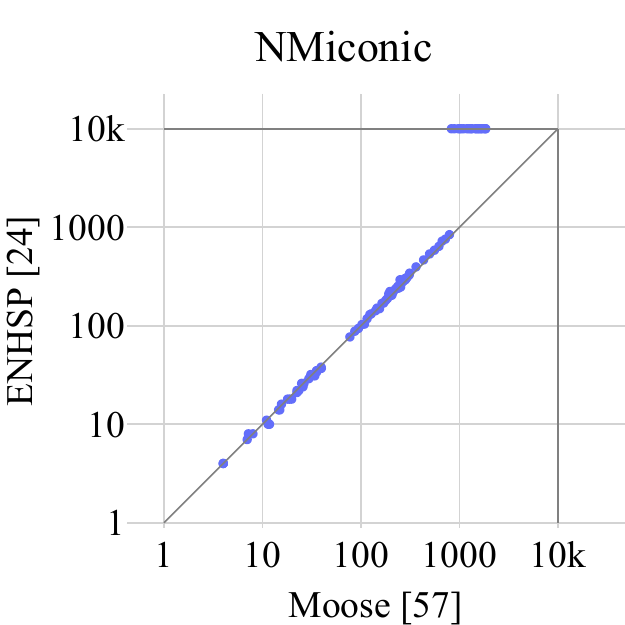}
  \includegraphics[width=\ccc\textwidth]{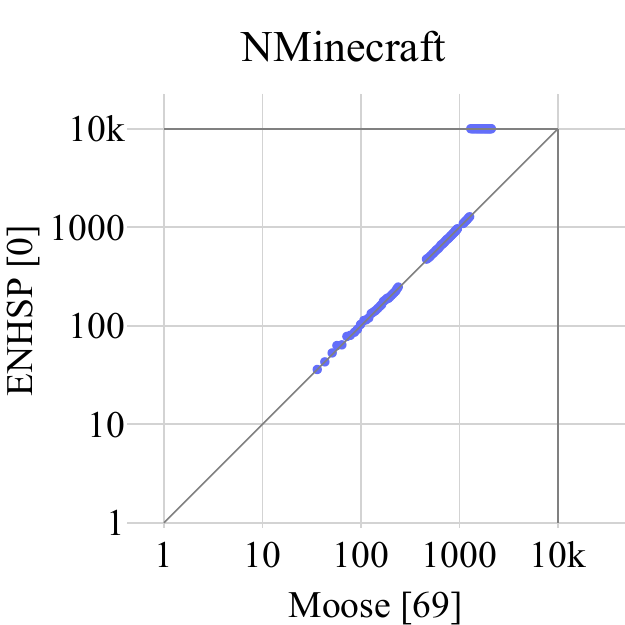}
  \includegraphics[width=\ccc\textwidth]{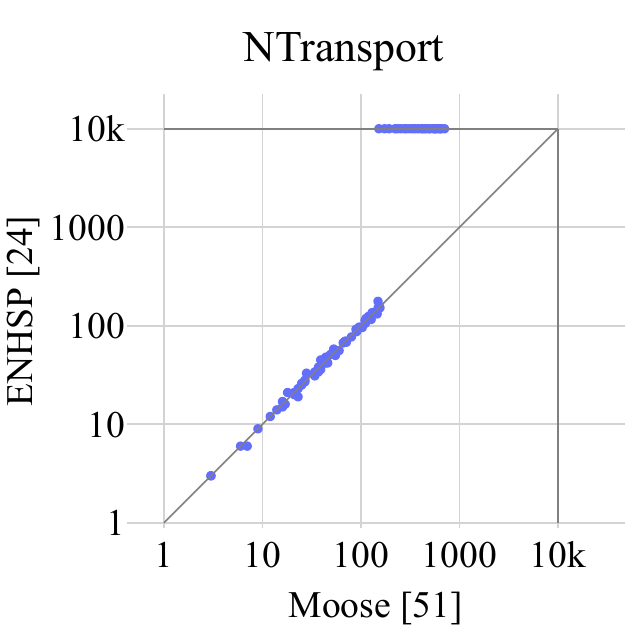}
  \caption{
    Plot comparisons of expanded nodes of the \enhspHMRPHJ{} configuration of \enhsp{} ($y$-axis) and \moose{} ($x$-axis) for different classical planning domains.
    A point $(x, y)$ represents the metric of the models indicated on the $x$ and $y$ axis on the domain.
    The number in the brackets next to each model indicates how many planning problems the model returned a higher quality plan than the model on the other axis.
    Points on the top left (resp.\ bottom right) triangle favour \moose{} (resp.\ \enhspHMRPHJ{}).
  }
  \label{fig:versus-enhsp}
\end{figure}

\section{Conjunctive Goal Regression}\label{app:sec:conjuncts}
In this section, we generalise the \moose{} algorithms by regressing over subsets of goals rather than singleton goals for synthesising more rules.
We then conduct experiments and analyse the effect of regressing over subsets of goals on the 3 metrics of synthesis cost, instantiation cost, and plan quality.

\subsection{Generalised \moose{} Algorithms}
The updated synthesis algorithm is presented in \Cref{alg:learn-k} which modifies \Cref{alg:learn}.
The changes are highlighted in blue and lie in \Crefrange{line:learn-k:a}{line:learn-k:b} and \Cref{line:learn-k:c}.
The subroutine $\fFont{extractRules'}$ is the same as $\fFont{extractRules}$ in \Cref{alg:extract-rules} with the changes where
\begin{enumerate}
  \item we modify the type of the precedence ranking function from $\rules \to \N$ to $\rules \to \N \times \N$ and \Cref{line:extract:precedence} is then changed to $\pi \la \pi \cup \set{(r, (-\abs{g}, \abs{\plan} - i + 1))}$, and
  \item rules where the state condition and goal condition have a non-empty intersection are pruned.
\end{enumerate}

Furthermore, \Cref{line:plan:query1} in \Cref{alg:plan} now iterates over $r \in \solution$ in ascending precedence values in lexicographical order.
For example, we would have $(-2, 3)$ queried before $(-2, 4)$ which in turn is queried before $(-1, 2)$.
The intuition here uses the triangle inequality where optimal plans for subgoals may be higher quality than achieving individual goals optimally in sequence.
Note however that this is not always the case, as two goals may be conflicting when trying to achieve them both as opposed to achieving each goal individually.
An example is trying to place 2 objects in a box but the box can only fit one object.

\begin{algorithm}[h!]
  \footnotesize
  \DontPrintSemicolon

  \caption{\moose{} Program Synthesis via $k$-Subset Goal Regression}
  \label{alg:learn-k}
  \KwInput{
    Training problems $\trainProblems = \problem^{(1)}, \ldots, \problem^{(n_t)}$, size of regressed goals $n_r$, and number of goal permutations $n_p \in \N$ (default: 3).
  }
  \KwOutput{
    \moose{} program $\solution$.
  }
  $\solution \la \emptyset$ \\
  \For{$i=1, \ldots, n_t$}{
    $n_g \la |\problem^{(i)}[g]|$ \\
    \For{$j=1, \ldots, \min(n_p, n_g!)$}{
      $s \la \problem^{(i)}[s_0]$ ; 
      $\vec{g} \la \fFont{newPermutation}(\problem^{(i)}[g])$ \\
      \color{blue}\For{$l=1, \ldots, n_r$}{\color{black} \label{line:learn-k:a}
        {\color{blue}$u \la \text{largest integer of the form $1 + k \cdot l$ such that $u \leq n_g - k + 1$}$} \\
        \color{blue}\For{$k=1, 1 + n_r, \ldots, u$}{\color{black}
          {\color{blue}$g' \la \set{\vec{g}_k, \ldots, \vec{g}_{\min(k + n_r - 1, n_g)}}$ \label{line:learn-k:b}} \\
          $\plan \la \fFont{optimalPlan}(\problem^{(i)}_{s, g'})$ \\
          \lIf{$\plan = \bot$}{
            \textbf{continue}
          }
          {\color{blue} $\solution \la \solution \cup \fFont{extractRules'}(\plan, g')$} \label{line:learn-k:c} \\
          $s \la \succ(s, \plan)$ \\
        }
      }
    }
  }
  \Return{$\solution$}
\end{algorithm}

\subsection{Experiments}
We perform experiments on the classical planning benchmarks with $n_r \in \set{1, 2, 3}$, where $n_r=1$ corresponds to the algorithms presented in the main paper, over 5 repeats.
Ferry results for $n_r = 3$ are omitted as no training problem exhibited more than 2 goal atoms.
We report results concerning the 3 metrics of synthesis cost, instantiation cost, and solution quality.

\paragraph{Synthesis Cost}
Synthesis costs are summarised in \Cref{tab:synthesis-k}.
We observe that increasing $n_r$ results in higher synthesis costs and the size of the synthesised generalised plan (\# learned rules).

\begin{table}[h!]
  \small
  \begin{tabularx}{\textwidth}{l *{10}{Y}}
    \toprule
    & \multicolumn{3}{c}{Time (s)} & \multicolumn{3}{c}{Memory (MB)} & \multicolumn{3}{c}{\# Learned Rules} \\
    \cmidrule(l){2-4}
    \cmidrule(l){5-7}
    \cmidrule(l){8-10}
    Domain & 1 & 2 & 3 & 1 & 2 & 3 & 1 & 2 & 3 \\
    \cmidrule(l){1-1}
    \cmidrule(l){2-4}
    \cmidrule(l){5-7}
    \cmidrule(l){8-10}
    Barman & 202 & 349 & 468 & 184 & 160 & 164 & 167 & 224 & 222 \\
    Ferry & 9 & 14 & $-$ & 52 & 59 & $-$ & 5 & 37 & $-$ \\
    Gripper & 10 & 10 & 12 & 64 & 65 & 52 & 4 & 10 & 19 \\
    Logistics & 71 & 113 & 159 & 73 & 113 & 148 & 90 & 980 & 2187 \\
    Miconic & 12 & 44 & 26 & 52 & 150 & 54 & 11 & 62 & 79 \\
    Rovers & 534 & 884 & 1061 & 187 & 209 & 317 & 88 & 2450 & 8481 \\
    Satellite & 514 & 740 & 951 & 82 & 142 & 194 & 13 & 252 & 1309 \\
    Transport & 21 & 31 & 43 & 80 & 84 & 59 & 5 & 82 & 184 \\
    \bottomrule
  \end{tabularx}
  \caption{Average runtime and memory usage ($\downarrow$) for synthesis, and number of learned rules over different number of $n_r \in \set{1, 2, 3}$ in \Cref{alg:learn-k}. Ferry results for $n_r = 3$ are omitted as no training problem exhibited more than 2 goal atoms.}
  \label{tab:synthesis-k}
\end{table}

\begin{figure}[h!]
  \centering
  \newcommand{\ccc}{0.225}
  \includegraphics[width=\ccc\textwidth]{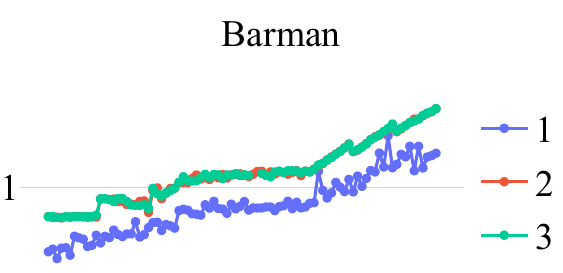}
  \includegraphics[width=\ccc\textwidth]{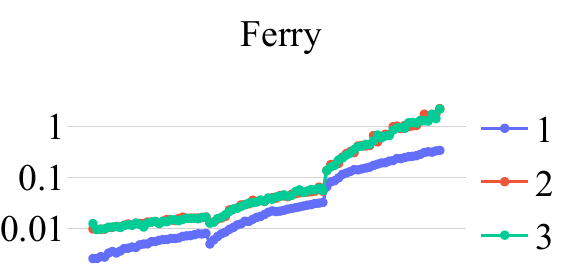}
  \includegraphics[width=\ccc\textwidth]{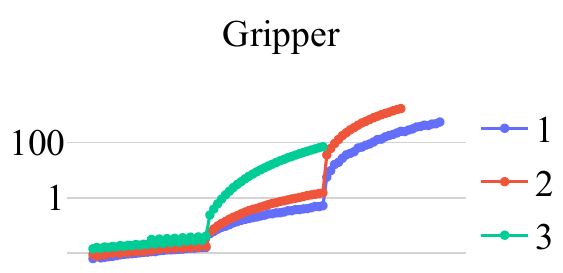}
  \includegraphics[width=\ccc\textwidth]{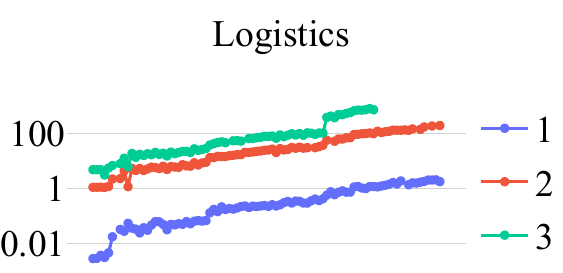}
  \includegraphics[width=\ccc\textwidth]{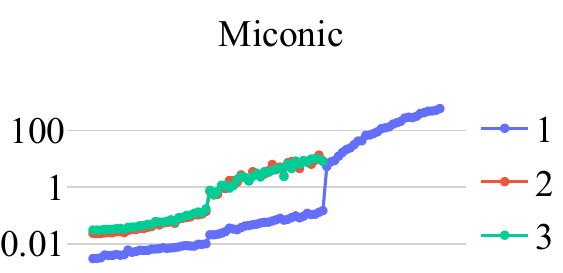}
  \includegraphics[width=\ccc\textwidth]{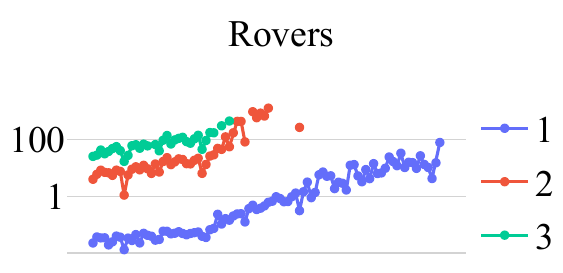}
  \includegraphics[width=\ccc\textwidth]{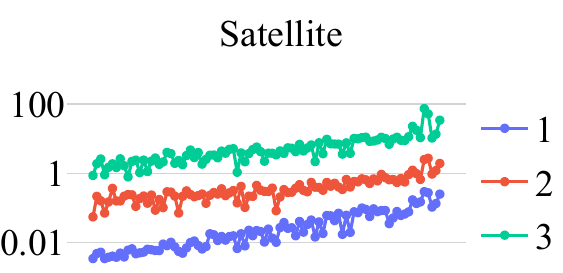}
  \includegraphics[width=\ccc\textwidth]{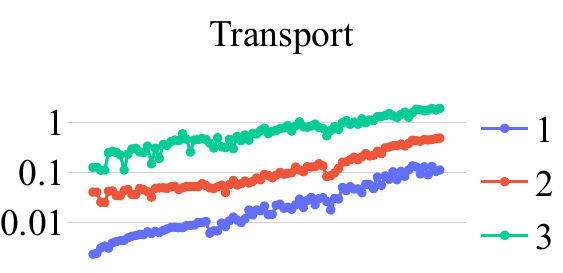}
  \caption{
    Average \textbf{time} in seconds in log scale ($\downarrow$) of different $n_r$ configurations across solved problems over different domains.
    Planning problem difficulty increases across the $x$-axis.
    Only points for which all seeds solve the problem are displayed.
    \label{fig:instantiation-k}
  }
\end{figure}

\paragraph{Instantiation Cost}
Instantiation costs are illustrated in \Cref{fig:instantiation-k}.
We observe that increasing $n_r$ results in higher instantiation costs, sometimes by more than an order of magnitude for each $n_r$ increment.
This is due to the increase in the number of synthesised rules and the associated costs of querying such rules.

\paragraph{Solution Quality}
Solution quality is aggregated in \Cref{tab:plan-length-k} by summing the plan length over problems that are solved by each $n_r$ configuration.
We observe that in all domains except Barman that increasing $n_r$ above 1 improves solution quality.
However, there is minimal change between $n_r=2$ and $n_r=3$.

\begin{table}[h!]
  \small
  \begin{tabularx}{\textwidth}{l Y Y r Y r}
    \toprule
    Domain & 1 & & 2 & & 3 \\
    \midrule
    Barman & 118778 & 121153 & $-2.0\%$ & 121167 & $-2.0\%$\\
    Ferry & 77760 & 70252 & $+9.7\%$ & 70247 & $+9.7\%$\\
    Gripper & 70800 & 53100 & $+25.0\%$ & 53190 & $+24.9\%$\\
    Logistics & 18224 & 18017 & $+1.1\%$ & 17967 & $+1.4\%$\\
    Miconic & 6293 & 6158 & $+2.1\%$ & 6172 & $+1.9\%$\\
    Rovers & 2027 & 1902 & $+6.2\%$ & 1888 & $+6.9\%$\\
    Satellite & 21923 & 18080 & $+17.5\%$ & 17862 & $+18.5\%$\\
    Transport & 14672 & 13586 & $+7.4\%$ & 13671 & $+6.8\%$\\
    \bottomrule
  \end{tabularx}
  \caption{Average \textbf{plan length} ($\downarrow$) of different $n_r$ configurations across solved problems over different domains. The percentages under the $n_r=2$ and $n_r=3$ columns indicate performance improvement relative to $n_r=1$.}
  \label{tab:plan-length-k}
\end{table}

\end{document}